\documentclass[sigconf]{acmart}

\setcopyright{none} % Removes the copyright statement
\renewcommand\footnotetextcopyrightpermission[1]{} % Removes the footnote with conference information
\acmConference[EvalMG @SIGIR '26]{Proceedings of the Second Workshop of Evaluation for Multi-Modal Generation}{July 24, 2026}{Melbourne, VIC, Australia}

\AtBeginDocument{%
  }

\usepackage{booktabs}
\usepackage{multirow}
\usepackage{rotating}
\usepackage{longtable}
\usepackage{cleveref}


\begin{document}

%%
%% Title.  Optional first argument is a short title for page headers.
\title[CLIP-CC-Bench]{CLIP-CC-Bench: Evaluating Paragraph-Level Video
  Descriptions in Video--Language Models}

%% Authors.
\author{Mukhtiar Ali}
\authornote{Equal contribution.}
\email{mukhtiar.ali@jacks.sdstate.edu}
\affiliation{%
  \institution{South Dakota State University}
  \city{Brookings}
  \state{SD}
  \country{USA}
}

\author{Harsh Dubey}
\authornotemark[1]
\email{harsh.dubey@jacks.sdstate.edu}
\affiliation{%
  \institution{South Dakota State University}
  \city{Brookings}
  \state{SD}
  \country{USA}
}

\author{Sugam Mishra}
\authornotemark[1]
\email{sugam.mishra@jacks.sdstate.edu}
\affiliation{%
  \institution{South Dakota State University}
  \city{Brookings}
  \state{SD}
  \country{USA}
}

\author{Chulwoo Pack}
\authornote{Corresponding author.}
\email{chulwoo.pack@sdstate.edu}
\affiliation{%
  \institution{South Dakota State University}
  \city{Brookings}
  \state{SD}
  \country{USA}
}

\renewcommand{\shortauthors}{Ali et al.}

%% Abstract.
\begin{abstract}
Benchmarking video-language models has largely focused on short clips and single-sentence metrics, leaving open whether current systems can generate accurate long-form, paragraph-level descriptions. We introduce CLIP-CC-Bench, an evaluation suite for long-form video description built from 5 hours of movie content segmented into 90-second clips, each paired with an expert-written paragraph-style reference. The evaluation suite employs an ensemble of five state-of-the-art LLM-based embedding models to increase reliability and mitigate single-model bias, and applies two complementary methodologies: (i) coarse-grained semantic matching and (ii) fine-grained semantic matching; to compare model-generated descriptions against CLIP-CC-Bench references. Using this framework, we evaluate 17 state-of-the-art video-language models and report both their Borda-aggregated rankings and their average scores on CLIP-CC-Bench. We further quantify the protocol's internal reliability through inter-judge agreement and bootstrap ranking stability. We release standardized evaluation scripts, model outputs, and aggregation tools at \url{https://github.com/Multimodal-Intelligence-Lab/CLIP-CC-Bench} to support reproducibility. CLIP-CC-Bench provides a practical evaluation framework for long-form video description, filling a gap left by existing short-clip and QA-only benchmarks.

\end{abstract}

%%
%% \maketitle processes the title/author/abstract/CCS/keywords block.
\maketitle

%%
%% Body sections.
\section{Introduction}
\label{sec:intro}

The rapid advancement of Video Language Models (VLMs) has demonstrated remarkable capabilities in understanding and describing visual content, particularly in video analysis tasks. Current state-of-the-art models such as VideoLLaMA3~\cite{zhang2025videollama3}, InternVL~\cite{chen2024internvl}, and mPLUG-Owl3~\cite{ye2024mplugowl3} have achieved impressive performance across various video understanding benchmarks. However, evaluating the quality of long-form, paragraph-level video descriptions generated by these models remains a significant challenge, particularly when assessing their ability to capture the overarching story and fine-grained details.

Traditional evaluation methods for video description tasks have predominantly relied on surface-level textual similarity metrics such as BLEU~\cite{papineni2002bleu}, ROUGE~\cite{lin2004rouge}, and METEOR~\cite{banerjee2005meteor}. While these metrics provide useful insights into lexical overlap between generated and reference descriptions, they often fail to capture the semantic understanding and discourse structure that are crucial for comprehensive video understanding. To move beyond purely local n-gram matching, CIDEr~\cite{vedantam2015cider} and SODA~\cite{fujita2020soda} have been proposed to place greater emphasis on informative content and event structure rather than raw lexical overlap. However, CIDEr remains fundamentally anchored to n-gram statistics and may still underrepresent deeper semantic and temporal coherence, while SODA relies on intermediate event representations with temporal boundaries and employs intersection-over-union calculations between predicted and reference timestamps. This dependency limits SODA's applicability to realistic long-form videos where multiple actions occur within brief temporal windows. Furthermore, SODA uses METEOR for event description similarity, which remains constrained to n-gram matching, and does not assess story-level comprehension across the entire video narrative. Recent advances in evaluation methodologies have introduced embedding-based similarity measures~\cite{zhang2019bertscore,lee2024nv} and LLM-as-judge frameworks~\cite{liu2023g}; yet LLM-as-judge approaches are typically black-box and exhibit significant consistency issues, with their scores showing limited or unstable correlation with human judgments, particularly for long-form video descriptions. Comprehensive benchmarks that systematically address these evaluation challenges for long-form video description remain limited.

Current video-language benchmarks present several fundamental limitations for evaluating long-form, paragraph-level video descriptions. Single-sentence captioning datasets like MSVD~\cite{chen2011msvd}, MSR-VTT~\cite{xu2016msr}, and VATEX~\cite{wang2019vatex} focus on describing isolated short clips without requiring comprehensive multi-event understanding. Dense captioning benchmarks such as ActivityNet Captions~\cite{krishna2017dense} and YouCook2~\cite{zhou2018youcook2} provide multi-sentence descriptions but evaluate segments independently, failing to assess holistic video understanding. Video QA benchmarks like TVQA~\cite{lei2018tvqa}, NExT-QA~\cite{xiao2021next}, and Video-MME~\cite{fu2024videomme} test discrete reasoning questions rather than the ability to generate comprehensive descriptions. Furthermore, many datasets include proper nouns and specific cultural references that can lead to spurious correlations and may not accurately reflect a model's fundamental video understanding capabilities.

To address these limitations, we introduce CLIP-CC-Bench, a comprehensive evaluation suite for long-form, paragraph-level video description. Our benchmark comprises of 5 hours of movie content segmented into 200 carefully selected 90-second movie clips with expert-written paragraph descriptions that systematically exclude proper nouns and cultural references, ensuring evaluation focuses on fundamental visual understanding rather than memorized associations. We make three key contributions:

\paragraph{Curated bias-minimizing dataset.} Our clips span diverse cinematographic styles and temporal dynamics, providing challenging evaluation scenarios while eliminating spurious correlations from proper nouns and specific cultural references.

\paragraph{Ensemble-based evaluation framework.} We employ an ensemble of five state-of-the-art LLM-based embedding models from the MTEB~\cite{muennighoff2023mteb} leaderboard to mitigate single-model bias and enhance evaluation reliability. Our framework applies two complementary methodologies: coarse-grained semantic matching for holistic paragraph-level alignment and fine-grained semantic matching for detailed content verification. Beyond a straightforward combination of existing similarity measures, the framework is designed as a structured multi-granular measurement protocol: the coarse--fine decomposition separates narrative-level alignment from detail coverage and exposes a systematic gap between the two, while rank-based Borda aggregation cancels the scale biases of individual embedding judges. We quantify the protocol's internal reliability directly---via inter-judge agreement and bootstrap stability of the ranking (\cref{sec:reliability})---and evaluate 17 state-of-the-art VLMs, reporting Borda-aggregated rankings and average scores.

\paragraph{Public benchmark release.} We provide all data, evaluation code, and baseline results at \url{https://github.com/Multimodal-Intelligence-Lab/CLIP-CC-Bench}, together with a project page and leaderboard at \url{https://multimodal-intelligence-lab.github.io/CLIP-CC-Bench/}, to facilitate reproducible research and continued advancement in video understanding evaluation.

\section{Related Work}
\label{sec:related}

\subsection{Video Description Benchmarks}

Single-sentence captioning datasets such as MSVD~\cite{chen2011msvd}, MSR-VTT~\cite{xu2016msr}, and VATEX~\cite{wang2019vatex} established foundational protocols for video--text alignment but restrict evaluation to isolated events in short clips. Dense captioning benchmarks extend temporal coverage---ActivityNet Captions~\cite{krishna2017dense} provides temporally localized event descriptions, while YouCook2~\cite{zhou2018youcook2} and TACoS~\cite{rohrbach2014tacos} target procedural activities---yet they evaluate segments independently with n-gram metrics, and even SODA~\cite{fujita2020soda}, which improves temporal alignment, still scores event descriptions with METEOR. Video QA benchmarks including TVQA~\cite{lei2018tvqa}, TGIF-QA~\cite{jang2017tgif}, NExT-QA~\cite{xiao2021next}, and DramaQA~\cite{choi2020drama} probe reasoning through question--answer pairs, and long-video suites such as Video-MME~\cite{fu2024videomme}, MVBench~\cite{li2023mvbench}, and LongVideoBench~\cite{wu2024longvideo} document substantial performance degradation as video length grows---but their multiple-choice format tests discrete recognition rather than the ability to generate coherent long-form descriptions. Specialized resources (Ego4D NLQ~\cite{grauman2022ego4d}, MovieGraphs~\cite{vicol2018moviegraphs}, M-VAD Names~\cite{pini2019mvad}) target temporal localization, scene graphs, or character naming rather than description quality. Across these families, no benchmark offers holistic assessment of paragraph-level description for minute-scale videos---the gap CLIP-CC-Bench addresses.

\subsection{Evaluation Methodologies for Video Description}

\paragraph{N-gram metrics.} BLEU~\cite{papineni2002bleu}, ROUGE~\cite{lin2004rouge}, METEOR~\cite{banerjee2005meteor}, and CIDEr~\cite{vedantam2015cider} measure surface-level lexical overlap. They provide standardized, reproducible protocols, but for paragraph-length descriptions they systematically miss semantically equivalent paraphrases and capture neither discourse structure nor narrative coverage.

\paragraph{Embedding-based similarity.} BERTScore~\cite{zhang2019bertscore} and Sentence-BERT~\cite{reimers2019sentence} capture semantic similarity beyond exact lexical matches, but their short context windows (typically 512 tokens) make them unsuitable for paragraph-level comparison. Recent LLM-based embedding models combine deep language understanding with context windows that accommodate full paragraphs. The MTEB benchmark~\cite{muennighoff2023mteb} ranks such models across clustering, semantic textual similarity, classification, and retrieval tasks; current leaders include NV-Embed-v2~\cite{lee2024nv}, KaLM-Embedding-Gemma3-12B~\cite{kalm2024}, Llama-Embed-Nemotron-8B~\cite{nvidia2024nemotron}, Qwen3-Embedding-8B~\cite{qwen2024embedding}, and GTE-Qwen2-7B-instruct~\cite{li2024gte}. These models excel at paragraph-level semantic matching, though fine-grained detail sensitivity remains challenging. EMScore~\cite{shi2022emscore} evaluates video captions through coarse- and fine-grained embedding matching, but operates cross-modally against the video and targets short captions rather than long-form, paragraph-level descriptions.

\paragraph{LLM-as-judge.} G-Eval~\cite{liu2023g} pioneered rubric-based LLM evaluation with chain-of-thought scoring, and G-VEval~\cite{tong2024gveval} extends this to video captions via GPT-4o, achieving strong correlation with human judgments on short captions. However, these frameworks remain largely unvalidated for long-form descriptions and inherit black-box limitations: opaque scoring rationales and high variance across repeated runs, which complicate systematic benchmarking.

\subsection{Video Language Models}

Building on early video--text pretraining (VideoBERT~\cite{sun2019videobert}, CBT~\cite{sun2019contrastive}), current VLMs span diverse designs: VideoLLaMA3~\cite{zhang2025videollama3} emphasizes temporal modeling, InternVL~\cite{chen2024internvl} unifies image and video understanding, MiniCPM-V~\cite{yao2024minicpm} targets parameter efficiency, mPLUG-Owl3~\cite{ye2024mplugowl3} uses multi-granular visual representations, LongVU~\cite{xu2024longvu} addresses extended video contexts, and ShareGPT4Video~\cite{chen2024sharegpt4video} and VideoChat~\cite{li2023videochat} focus on conversational understanding. Systematic evaluation of their long-form description capabilities, however, remains limited.

In summary, existing practice either relies on single n-gram metrics that miss semantic equivalence, on short-context embedding methods unsuited to paragraphs, or on LLM judges with transparency and consistency concerns. CLIP-CC-Bench instead employs an ensemble of five long-context MTEB embedding models with complementary coarse-grained (paragraph-level) and fine-grained (sentence-level) semantic matching, providing a transparent and reproducible protocol for paragraph-level video description evaluation.

\section{Methodology}
\label{sec:methodology}

This section details the construction of CLIP-CC-Bench, a carefully curated dataset designed to evaluate long-form, paragraph-level video description capabilities.

\begin{figure*}[t]
\centering
\includegraphics[width=\textwidth]{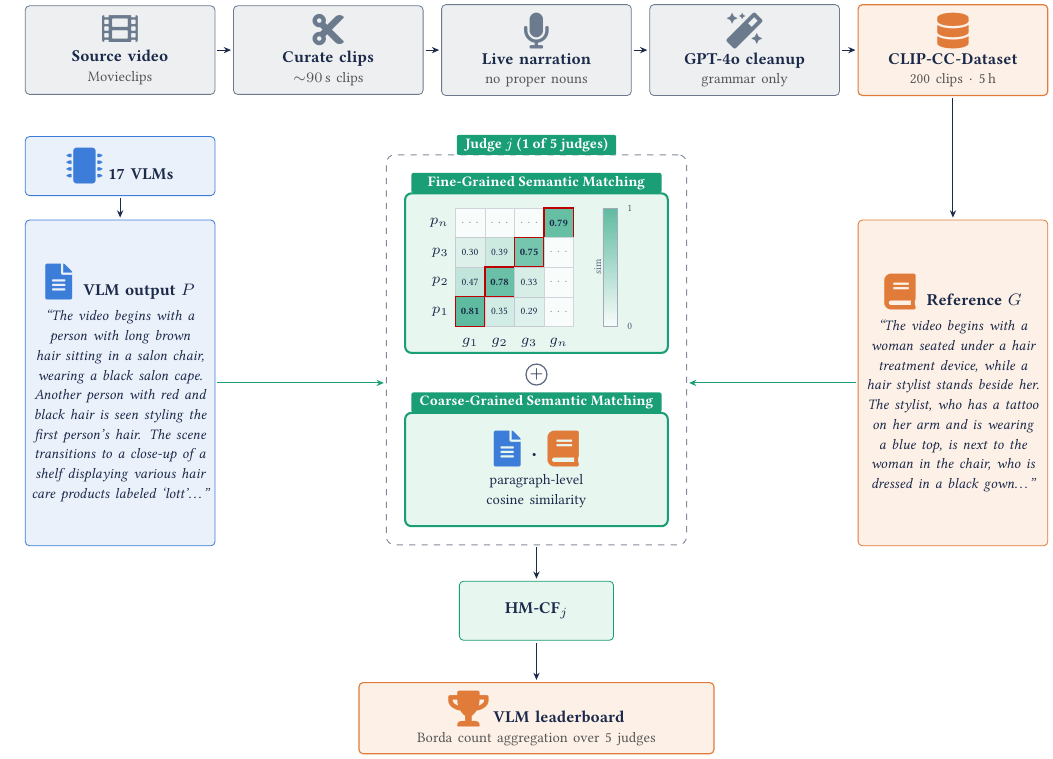}
\caption{Overview of CLIP-CC-Bench. Top: dataset construction curates $\sim$90\,s movie clips, collects live narration without proper nouns, applies a grammar-only GPT-4o cleanup, and yields the CLIP-CC-Dataset (200 clip--paragraph pairs, 5\,h total). Bottom: per-judge evaluation. A candidate paragraph $P$ from one of 17 VLMs (left; sample shown is mPLUG-Owl3 on clip~092) and the reference paragraph $G$ from the dataset (right) are scored by Judge $j$, one of five MTEB embedding judges. The judge combines an $n{\times}n$ sentence-level cosine matrix (Fine-Grained Semantic Matching; greedy diagonal best-matches outlined in red) with the paragraph-level cosine (Coarse-Grained Semantic Matching) into a harmonic-mean $\mathrm{HM\text{-}CF}_{j}$. Borda aggregation across the five judges produces the leaderboard over the 17 VLMs.}
\Description{Two-part overview figure. Top: a left-to-right pipeline of five icon-labelled boxes for dataset construction --- Source video (Movieclips), Curate clips (~90s), Live narration (no proper nouns), GPT-4o cleanup (grammar only), and CLIP-CC-Dataset (200 clips, 5h). Bottom: an evaluation diagram. On the left, a pink box labelled 17 VLMs sits above a tall pink box labelled VLM output P containing a sample paragraph beginning ``The video begins with a person with long brown hair...''. On the right, the CLIP-CC-Dataset box feeds a tall orange box labelled Reference G with a paragraph beginning ``The video begins with a woman seated under a hair treatment device...''. Both side boxes feed arrows into a central dashed container labelled Judge j (1 of 5 judges), which holds two stacked compartments joined by a circled plus. The upper compartment, Fine-Grained Semantic Matching, shows a 4 by 4 cosine matrix with rows p1, p2, p3, pn and columns g1, g2, g3, gn; the n-th row and column are ellipsis cells; four diagonal cells (p1-g1, p2-g2, p3-g3, pn-gn) are outlined in red with values 0.81, 0.78, 0.75, 0.79; a vertical teal color-scale legend labelled sim runs from 0 to 1 on the right. The lower compartment, Coarse-Grained Semantic Matching, contains a document icon and a book icon joined by a dot, captioned ``paragraph-level cosine similarity''. Below the container, a box labelled HM-CF subscript j flows into a final orange box labelled VLM leaderboard, captioned ``Borda count aggregation over 5 judges''.}
\label{fig:overview}
\end{figure*}

\subsection{Dataset Construction Principles}

Our dataset construction methodology prioritizes temporal diversity and visual richness while systematically eliminating potential confounds from proper nouns and memorized associations. These principles ensure that CLIP-CC-Bench evaluates fundamental video understanding capabilities for long-form, paragraph-level description generation.

\subsubsection{Temporal Scope and Content Selection}
Our video content selection emphasizes temporal and visual diversity to provide comprehensive evaluation of VLM capabilities across different cinematographic styles and narrative structures. We curate content that differs significantly from contemporary training data patterns, ensuring models are challenged beyond their typical training distribution through diverse visual effects techniques, color palettes, and production methodologies.

The temporal scope of our dataset ensures sufficient distance from recent internet content to minimize potential data contamination issues. Given that most contemporary VLMs are trained on internet-scraped data heavily biased toward recent content, our careful curation reduces the likelihood that models have encountered similar visual patterns during training.

Our content curation process focuses exclusively on movie clips that contain complex scenes with multiple actors, dynamic camera movements, and rich visual storytelling elements. We prioritize clips that demonstrate temporal progression and character interactions that require sophisticated understanding to describe accurately.

\subsubsection{Proper Noun Elimination Strategy}
A fundamental design principle of CLIP-CC-Bench is the systematic elimination of proper nouns from all ground-truth descriptions. This design choice addresses a critical limitation in existing benchmarks, where models may leverage memorized associations rather than demonstrating genuine visual understanding.

Our annotation guidelines explicitly prohibit the use of:
\begin{itemize}
    \item Character names and celebrity identifications
    \item Specific geographical locations and landmarks
    \item Brand names and commercial references
    \item Cultural or historical proper nouns
    \item Fictional universe-specific terminology
\end{itemize}

Instead, annotators employ descriptive alternatives such as \textit{``a man in a black jacket''} rather than character names, or \textit{``a luxury sedan''} instead of specific car models. This approach ensures that evaluation focuses on fundamental visual understanding capabilities rather than world knowledge retrieval.

\subsection{Video Selection and Processing}

We implement rigorous selection and processing protocols to ensure consistent quality and comprehensive coverage across our evaluation suite. Our systematic approach balances visual complexity with evaluation standardization requirements for long-form video description assessment.

\subsubsection{Content Criteria}
Our video selection process employs rigorous criteria to ensure benchmark quality and diversity. Each selected clip must satisfy the following requirements:

\paragraph{Duration Standardization.} All clips are standardized to approximately 90 seconds to ensure consistent evaluation conditions while providing sufficient content for comprehensive description.

\paragraph{Visual Complexity.} Selected scenes must contain multiple visual elements including character interactions, environmental details, and temporal dynamics that require sophisticated understanding to describe accurately.

\paragraph{Self-Contained Content.} Clips must represent self-contained segments that can be understood without external context, ensuring fair evaluation across different models.

\paragraph{Technical Quality.} All video content maintains consistent resolution and audio quality standards to prevent technical artifacts from influencing model performance.

\subsection{Annotation Procedure}
\label{sec:annotation}

\paragraph{Annotators.} The reference descriptions were written by four graduate-student annotators fluent in English, all trained on the proper-noun-exclusion guidelines above before annotating.

\paragraph{Procedure.} Each clip was assigned to a single annotator, who watched the $\sim$90\,s segment and narrated its description aloud in real time while deliberately avoiding proper nouns and other memorizable references. Each narration was transcribed by automatic speech recognition and then manually verified against the audio, and finally passed through a grammar-only GPT-4o cleanup that fixed disfluencies and grammar without adding, removing, or reordering content or introducing names (the cleanup prompt is given in \cref{sec:appendix:prompt}). Annotation took roughly 45 minutes per clip ($\sim$30\,min narration, $\sim$15\,min review), or about 150 hours in total.

\subsection{Dataset Statistics and Characteristics}

The final CLIP-CC-Bench dataset comprises 5 hours of movie content broken down into 90-second clips with corresponding expert-generated descriptions. Key statistics are presented in \cref{tab:dataset_stats}.

\begin{table}[t]
\centering
\caption{CLIP-CC-Bench dataset statistics.}
\label{tab:dataset_stats}
\begin{tabular}{lc}
\toprule
\textbf{Metric} & \textbf{Value} \\
\midrule
Number of videos & 200 \\
Clip duration & $\sim$90 seconds \\
Total video content & 5 hours \\
Distinct source films/series & $>$140 \\
Source release years & 1959--2024 \\
Words per description (mean $\pm$ std) & 402.2 $\pm$ 208.5 \\
Words per description [min, max] & [99, 1011] \\
Sentences per description (mean $\pm$ std) & 21.9 $\pm$ 11.5 \\
Vocabulary size (unique words) & 4,000 \\
\bottomrule
\end{tabular}
\end{table}

\begin{figure}[t]
\centering
\includegraphics[width=0.48\textwidth]{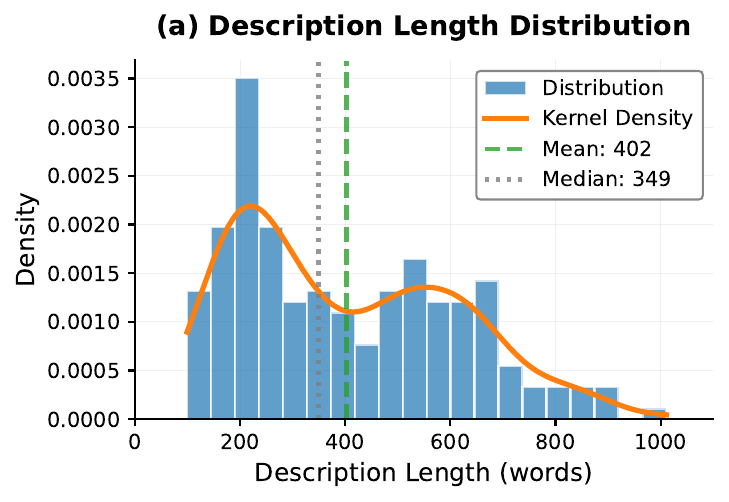}
\caption{Distribution of description characteristics in CLIP-CC-Bench. The dataset exhibits substantial variation in both word count and sentence length, reflecting the diverse complexity of video content while maintaining comprehensive paragraph-level descriptions suitable for evaluating long-form video understanding capabilities.}
\Description{Two side-by-side histograms summarizing per-clip description statistics across the 200 CLIP-CC-Bench videos. The left panel plots word count per description on the horizontal axis (range roughly 100 to 1000) and shows a bimodal shape, with one mode near 200--300 words and a second, broader mode around 400--800 words. The right panel plots sentence count per description (range roughly 5 to 60) and shows a single, right-skewed peak centered near 22 sentences, with a long tail toward higher counts.}
\label{fig:dataset_distribution}
\end{figure}

\Cref{fig:dataset_distribution} illustrates the distribution of description lengths across our dataset, demonstrating the substantial variation in complexity that reflects the diverse nature of video content. The bimodal distribution in word counts reveals two primary description patterns: concise summaries for simpler scenes (100-300 words) and comprehensive narratives for complex sequences (400-800 words). This natural variation ensures that CLIP-CC-Bench evaluates models across a spectrum of descriptive requirements, from brief action sequences to intricate multi-character interactions.

\paragraph{Source and Topical Diversity.} The 200 clips are drawn from more than 140 distinct films and series spanning 1959--2024 (median release year 2009): 110 sources contribute exactly one clip, and no single source contributes more than six clips (3\%). Categorizing clips by primary narrative type yields four categories with at least 5\% representation---Dialogue/Drama (52\%), Violence/Combat (13\%), Action/Chase (11.5\%), and Public/Social (11.5\%)---with the remainder spanning medical/procedural, dramatic-confrontation, crime, domestic, suspense, and romance scenes (full distribution in \cref{tab:appendix:topics}). Unsupervised clustering of CLIP video embeddings corroborates this spread independently of the category labels: the largest visual cluster contains only 17\% of clips, and seven clusters each hold at least 5\%. The dialogue-heavy distribution is expected for narrative film content, while the remaining 48\% of clips provide dense action, multi-character, and procedural challenges.

Our ground-truth descriptions exhibit several distinctive characteristics that differentiate CLIP-CC-Bench from existing datasets:

\paragraph{Comprehensive Detail.} Descriptions provide thorough coverage of visual content with substantial variation in length to accommodate different scene complexities.

\paragraph{Temporal Structure.} Descriptions follow chronological progression through the video content, explicitly tracking temporal relationships between events and character actions.

\paragraph{Visual Specificity.} Detailed descriptions of visual elements including clothing, environmental features, object properties, and spatial relationships provide rich evaluation targets for VLM assessment.

\paragraph{Action Coverage.} Both foreground actions (primary character activities) and background elements (environmental details, secondary characters) are systematically documented to enable comprehensive evaluation.

This composition ensures representative coverage of diverse cinematic styles while maintaining evaluation consistency and quality standards for long-form description assessment.

\section{Experiments}
\label{sec:experiments}

We evaluate 17 state-of-the-art Video Language Models on CLIP-CC-Bench using our ensemble-based multi-granular evaluation framework to analyze their long-form video description capabilities.

\subsection{Model Selection and Inference}

We evaluate 17 representative VLMs spanning different architectural paradigms and training methodologies, as summarized in \cref{tab:models}. To ensure comprehensive evaluation of each model's capabilities, we utilize their maximum supported frame capacity, allowing each VLM to process videos at its full potential rather than imposing artificial constraints.

\begin{table}[t]
\centering
\caption{Video Language Models evaluated on CLIP-CC-Bench.}
\label{tab:models}
\resizebox{\columnwidth}{!}{
\begin{tabular}{llcc}
\toprule
\textbf{Model} & \textbf{Architecture Family} & \textbf{Parameters} & \textbf{Max Frames} \\
\midrule
LLaVA-OneVision & Transformer & 7B & 8 \\
LLaVA-NeXT-Video & Transformer & 7B & 32 \\
VideoLLaMA3 & Transformer & 7B & 32 \\
InternVL2 & Transformer & 8B & 12 \\
Qwen2.5 & Transformer & 32B & -- \\
Qwen2.5 & Transformer & 72B & -- \\
\midrule
LongVU & Efficient & 7B & -- \\
LongVA & Efficient & 7B & 128 \\
MiniCPM-V & Efficient & 8B & 32 \\
\midrule
Video-XL & Specialized & 7B & 16 \\
TimeChat & Specialized & 7B & 16 \\
TS-LLaVA & Specialized & 7B & 32 \\
VideoChat-Flash & Specialized & 2B & 16 \\
\midrule
Oryx & Multimodal & 7B & -- \\
ViLAMP & Multimodal & 7B & 600 \\
mPLUG-Owl3 & Multimodal & 7B & 16 \\
ShareGPT4Video & Multimodal & 8B & -- \\
\bottomrule
\end{tabular}
}
\end{table}

All models generate paragraph-length descriptions for each video clip using standardized inference parameters while operating at their maximum frame processing capacity.\footnote{Model outputs are unavailable for one clip (\#126) across all 17 VLMs; all reported metrics are therefore computed over the identical remaining 199 clips, keeping comparisons strictly paired.}

\subsection{Evaluation Protocol}
We standardize evaluation with: (1) uniform prompting: \textit{``Provide a detailed description of the video, covering all significant events, the actions of each character or entity, any camera movements, the attributes of the characters or entities, and a description of the scene, focusing on the characters themselves without recognizing, identifying, or naming them, and only describing their appearance, behavior, and actions.''}, (2) consistent generation parameters (temperature=0.0, max\_tokens=5000), and (3) deterministic inference with greedy decoding to ensure reproducible and consistent outputs.

\subsection{Evaluation Methodologies}
\label{sec:eval_methods}

We employ a multi-granular semantic matching framework that evaluates VLM-generated descriptions through both holistic and sentence-level alignment with reference descriptions. This approach combines embedding-based similarity at multiple scales with a multi-judge consensus mechanism to ensure robust and comprehensive performance assessment.

\paragraph{Notation.}
Let $G = \{g_1, \dots, g_m\}$ denote a reference paragraph from CLIP-CC-Bench with $m$ sentences and $P = \{p_1, \dots, p_n\}$ denote a candidate paragraph generated by a VLM with $n$ sentences. We treat the five MTEB embedding models as independent judges $j \in \{1, \dots, J\}$ with $J{=}5$, where each judge $j$ realizes an encoder $f_j(\cdot)$ that maps a sentence or a whole paragraph to a dense vector. Cosine similarity between two vectors $u, v$ is $\cos(u, v) = \frac{u \cdot v}{\|u\|\,\|v\|}$.

\subsubsection{Coarse-Grained Semantic Matching}
We encode the complete reference and predicted paragraphs as single dense vector representations and measure their cosine similarity, capturing overall semantic coherence between the generated description and the reference:
\begin{equation}
\mathrm{Coarse}_j(G, P) \;=\; \cos\bigl(f_j(G),\, f_j(P)\bigr).
\label{eq:coarse}
\end{equation}

\subsubsection{Fine-Grained Semantic Matching}
After sentence-level tokenization, we embed each sentence individually and form an $n{\times}m$ cosine matrix between predicted and reference sentences. Fine-grained \emph{precision} averages, over predicted sentences, the cosine to their best-matching reference sentence; \emph{recall} averages the symmetric quantity from the reference side:
\begin{align}
\mathrm{Pr}_j(G, P) &\;=\; \frac{1}{n} \sum_{i=1}^{n} \max_{1 \le k \le m} \cos\bigl(f_j(p_i),\, f_j(g_k)\bigr), \label{eq:precision} \\
\mathrm{Rc}_j(G, P) &\;=\; \frac{1}{m} \sum_{k=1}^{m} \max_{1 \le i \le n} \cos\bigl(f_j(g_k),\, f_j(p_i)\bigr). \label{eq:recall}
\end{align}
The sentence-level F1 harmonically combines the two, balancing semantic precision and coverage:
\begin{equation}
\mathrm{Fine}_j(G, P) \;=\; \frac{2 \cdot \mathrm{Pr}_j(G, P) \cdot \mathrm{Rc}_j(G, P)}{\mathrm{Pr}_j(G, P) + \mathrm{Rc}_j(G, P)}.
\label{eq:fine_f1}
\end{equation}

\subsubsection{Multi-Judge Consensus Ranking}
For each judge $j$, we summarize a $(G, P)$ pair with the harmonic mean of its coarse and fine scores, producing a per-judge metric that penalizes models that excel in only one granularity:
\begin{equation}
\mathrm{HM\text{-}CF}_j(G, P) \;=\; \frac{2 \cdot \mathrm{Coarse}_j(G, P) \cdot \mathrm{Fine}_j(G, P)}{\mathrm{Coarse}_j(G, P) + \mathrm{Fine}_j(G, P)}.
\label{eq:hmcf}
\end{equation}
Each judge then ranks the $V{=}17$ VLMs by their dataset-averaged $\mathrm{HM\text{-}CF}_j$. Letting $\mathrm{rank}_j(v) \in \{1, \dots, V\}$ denote the rank of VLM $v$ under judge $j$, the final consensus is obtained via Borda count:
\begin{equation}
\mathrm{Borda}(v) \;=\; \sum_{j=1}^{J} \bigl(V - \mathrm{rank}_j(v)\bigr),
\label{eq:borda}
\end{equation}
which assigns $V{-}1{=}16$ points to a first place and $0$ points to a last place under each judge, yielding a maximum attainable Borda score of $J(V{-}1) = 80$. This aggregation mitigates idiosyncratic biases of any single judge and provides a robust overall VLM ranking.

\section{Results}
\label{sec:results}

We present comprehensive evaluation results for 17 state-of-the-art Video Language Models on CLIP-CC-Bench using our ensemble-based evaluation framework. Our methodology employs five leading LLM-based embedding models from the MTEB~\cite{muennighoff2023mteb} leaderboard: GTE-Qwen2-7B~\cite{li2024gte}, KaLM-Embedding-Gemma3-12B~\cite{kalm2024}, Llama-Embed-Nemotron-8B~\cite{nvidia2024nemotron}, nv-embed-v2~\cite{lee2024nv}, and Qwen3-Embedding-8B~\cite{qwen2024embedding}. For each VLM, we compute coarse-grained semantic similarity (paragraph-level alignment) and fine-grained semantic similarity (sentence-level F1 matching) across all five embedding models. We then rank VLMs using Borda count based on their harmonic mean of coarse and fine scores ($\mathrm{HM\text{-}CF}$, see Eq.~\ref{eq:hmcf}) for each embedding model, and report mean $\mathrm{HM\text{-}CF}$ scores averaged across all five judges.

\subsection{Per-Judge Evaluation Statistics}

\Cref{tab:detailed_stats} presents detailed performance statistics for all 17 VLMs across the five embedding models. For each embedding model, we report coarse-grained similarity scores, fine-grained F1 scores, and their harmonic mean, along with standard deviations. The hierarchical structure reveals how different embedding models assess video description quality with varying sensitivities.

\begin{table*}[t]
\centering
\caption{Detailed evaluation statistics across five MTEB embedding models. For each model, we report Coarse (coarse-grained similarity, Eq.~\ref{eq:coarse}), Fine (fine-grained F1, Eq.~\ref{eq:fine_f1}), and HM-CF (harmonic mean of Coarse and Fine, Eq.~\ref{eq:hmcf}) as mean$\pm$std.}
\label{tab:detailed_stats}
\resizebox{\textwidth}{!}{
\begin{tabular}{lrrrrrrrrrrrrrrr}
\toprule
& \multicolumn{3}{c}{\textbf{GTE-Qwen2-7B}} & \multicolumn{3}{c}{\textbf{KaLM-Gemma3-12B}} & \multicolumn{3}{c}{\textbf{Nemo-8B}} & \multicolumn{3}{c}{\textbf{NV-Embed-v2}} & \multicolumn{3}{c}{\textbf{Qwen3-8B}} \\
\textbf{VLM} & \rotatebox{90}{\textbf{Coarse}} & \rotatebox{90}{\textbf{Fine}} & \rotatebox{90}{\textbf{HM-CF}} & \rotatebox{90}{\textbf{Coarse}} & \rotatebox{90}{\textbf{Fine}} & \rotatebox{90}{\textbf{HM-CF}} & \rotatebox{90}{\textbf{Coarse}} & \rotatebox{90}{\textbf{Fine}} & \rotatebox{90}{\textbf{HM-CF}} & \rotatebox{90}{\textbf{Coarse}} & \rotatebox{90}{\textbf{Fine}} & \rotatebox{90}{\textbf{HM-CF}} & \rotatebox{90}{\textbf{Coarse}} & \rotatebox{90}{\textbf{Fine}} & \rotatebox{90}{\textbf{HM-CF}} \\
\midrule
VideoLLaMA3 & 0.75±0.05 & 0.64±0.04 & 0.69±0.04 & 0.82±0.05 & 0.76±0.03 & 0.79±0.03 & 0.68±0.07 & 0.57±0.04 & 0.62±0.05 & 0.68±0.10 & 0.47±0.07 & 0.55±0.08 & 0.72±0.06 & 0.69±0.03 & 0.70±0.04 \\
mPLUG-Owl3 & 0.74±0.06 & 0.64±0.04 & 0.68±0.04 & 0.79±0.07 & 0.75±0.03 & 0.77±0.05 & 0.67±0.08 & 0.55±0.04 & 0.61±0.05 & 0.63±0.13 & 0.46±0.06 & 0.53±0.08 & 0.70±0.07 & 0.68±0.04 & 0.69±0.05 \\
ViLAMP & 0.71±0.07 & 0.62±0.05 & 0.66±0.05 & 0.78±0.07 & 0.75±0.03 & 0.76±0.04 & 0.65±0.08 & 0.55±0.05 & 0.59±0.06 & 0.60±0.14 & 0.44±0.06 & 0.50±0.08 & 0.69±0.07 & 0.66±0.04 & 0.68±0.05 \\
LLaVA-OneVision & 0.70±0.06 & 0.62±0.04 & 0.65±0.04 & 0.79±0.05 & 0.74±0.03 & 0.76±0.04 & 0.65±0.07 & 0.54±0.05 & 0.59±0.05 & 0.64±0.07 & 0.44±0.06 & 0.52±0.06 & 0.68±0.07 & 0.67±0.04 & 0.67±0.05 \\
LongVU & 0.70±0.06 & 0.60±0.05 & 0.64±0.05 & 0.78±0.06 & 0.73±0.03 & 0.75±0.04 & 0.63±0.08 & 0.51±0.05 & 0.56±0.06 & 0.61±0.10 & 0.44±0.06 & 0.51±0.07 & 0.67±0.07 & 0.67±0.03 & 0.67±0.05 \\
Qwen2.5-72B & 0.66±0.07 & 0.59±0.04 & 0.63±0.05 & 0.77±0.06 & 0.72±0.03 & 0.75±0.04 & 0.63±0.08 & 0.50±0.05 & 0.56±0.06 & 0.59±0.09 & 0.43±0.06 & 0.50±0.06 & 0.68±0.07 & 0.65±0.04 & 0.66±0.05 \\
Qwen2.5-32B & 0.66±0.06 & 0.57±0.04 & 0.61±0.04 & 0.77±0.05 & 0.72±0.03 & 0.74±0.03 & 0.60±0.07 & 0.49±0.04 & 0.54±0.05 & 0.58±0.08 & 0.42±0.05 & 0.49±0.06 & 0.66±0.06 & 0.65±0.03 & 0.65±0.04 \\
VideoChat-Flash & 0.63±0.07 & 0.59±0.04 & 0.61±0.05 & 0.74±0.06 & 0.72±0.03 & 0.73±0.04 & 0.59±0.09 & 0.52±0.05 & 0.55±0.06 & 0.54±0.11 & 0.40±0.06 & 0.45±0.08 & 0.67±0.06 & 0.64±0.04 & 0.66±0.05 \\
MiniCPM-V & 0.67±0.06 & 0.55±0.04 & 0.60±0.04 & 0.78±0.05 & 0.70±0.03 & 0.74±0.03 & 0.61±0.07 & 0.47±0.04 & 0.53±0.05 & 0.60±0.07 & 0.40±0.05 & 0.48±0.06 & 0.66±0.06 & 0.64±0.03 & 0.65±0.04 \\
Video-XL & 0.63±0.08 & 0.57±0.05 & 0.60±0.06 & 0.74±0.06 & 0.72±0.03 & 0.73±0.04 & 0.56±0.09 & 0.49±0.04 & 0.52±0.06 & 0.55±0.09 & 0.40±0.06 & 0.47±0.06 & 0.62±0.08 & 0.65±0.04 & 0.63±0.05 \\
ShareGPT4Video & 0.64±0.07 & 0.57±0.05 & 0.60±0.05 & 0.72±0.07 & 0.70±0.03 & 0.71±0.05 & 0.55±0.08 & 0.48±0.05 & 0.51±0.06 & 0.55±0.11 & 0.40±0.06 & 0.46±0.08 & 0.61±0.07 & 0.64±0.04 & 0.62±0.05 \\
InternVL2 & 0.63±0.09 & 0.56±0.06 & 0.60±0.08 & 0.74±0.07 & 0.71±0.04 & 0.72±0.05 & 0.56±0.10 & 0.49±0.05 & 0.52±0.07 & 0.57±0.10 & 0.38±0.06 & 0.45±0.07 & 0.61±0.08 & 0.63±0.05 & 0.62±0.06 \\
TimeChat & 0.58±0.07 & 0.55±0.04 & 0.57±0.05 & 0.72±0.06 & 0.71±0.03 & 0.71±0.04 & 0.52±0.08 & 0.49±0.05 & 0.50±0.06 & 0.53±0.09 & 0.35±0.06 & 0.42±0.07 & 0.59±0.07 & 0.62±0.04 & 0.60±0.05 \\
LLaVA-NeXT-Video & 0.55±0.09 & 0.53±0.05 & 0.54±0.06 & 0.69±0.07 & 0.71±0.03 & 0.70±0.05 & 0.49±0.10 & 0.48±0.05 & 0.48±0.07 & 0.50±0.10 & 0.37±0.07 & 0.42±0.08 & 0.56±0.09 & 0.63±0.04 & 0.59±0.06 \\
TS-LLaVA & 0.54±0.10 & 0.51±0.06 & 0.53±0.07 & 0.68±0.08 & 0.69±0.04 & 0.68±0.06 & 0.46±0.11 & 0.46±0.06 & 0.46±0.08 & 0.47±0.11 & 0.35±0.07 & 0.40±0.08 & 0.56±0.09 & 0.62±0.05 & 0.58±0.07 \\
Oryx & 0.52±0.10 & 0.51±0.05 & 0.51±0.07 & 0.67±0.07 & 0.68±0.04 & 0.67±0.05 & 0.47±0.10 & 0.45±0.05 & 0.46±0.07 & 0.48±0.10 & 0.34±0.07 & 0.40±0.08 & 0.56±0.09 & 0.61±0.05 & 0.58±0.07 \\
LongVA & 0.50±0.10 & 0.46±0.07 & 0.48±0.08 & 0.62±0.08 & 0.66±0.04 & 0.64±0.06 & 0.41±0.10 & 0.41±0.06 & 0.40±0.07 & 0.40±0.10 & 0.31±0.07 & 0.35±0.08 & 0.49±0.10 & 0.57±0.05 & 0.53±0.08 \\
\bottomrule
\end{tabular}
}
\end{table*}

Across all embedding models, VideoLLaMA3 consistently achieves the highest scores, with KaLM-Gemma3-12B showing particularly strong discrimination (HM-CF: 0.79). The standard deviations reveal model-specific variance patterns: NV-Embed-v2 exhibits the highest variability in coarse-grained scores (VideoLLaMA3: 0.10, mPLUG-Owl3: 0.13), while Llama-Embed-Nemotron-8B shows more conservative absolute scores but tighter distributions. This diversity across embedding models motivates our ensemble approach to reduce single-model bias.

\subsection{Overall VLM Ranking}

\Cref{tab:vlm_ranking} presents our final VLM ranking computed using Borda count aggregation across all five embedding models. For each embedding model, we rank VLMs by their $\mathrm{HM\text{-}CF}$ score and assign Borda points (16 points for rank 1, 15 for rank 2, down to 0 for rank 17). The final ranking is determined by total Borda score, with Mean (average $\mathrm{HM\text{-}CF}$ across all five embedding models) serving as a tiebreaker.

\begin{table}[t]
\centering
\caption{Overall VLM ranking on CLIP-CC-Bench using Borda count aggregation (Eq.~\ref{eq:borda}) across five MTEB embedding models. Borda scores represent cumulative ranking points across all judges. Mean is the average $\mathrm{HM\text{-}CF}$ across all five embedding models.}
\label{tab:vlm_ranking}
\begin{tabular}{clcc}
\toprule
\textbf{Rank} & \textbf{VLM} & \textbf{Borda} & \textbf{Mean} \\
\midrule
1 & VideoLLaMA3 & 80 & 0.67 \\
2 & mPLUG-Owl3 & 75 & 0.66 \\
3 & LLaVA-OneVision & 67 & 0.64 \\
4 & ViLAMP & 67 & 0.64 \\
5 & LongVU & 61 & 0.63 \\
6 & Qwen2.5-72B & 55 & 0.62 \\
7 & Qwen2.5-32B & 48 & 0.61 \\
8 & VideoChat-Flash & 42 & 0.60 \\
9 & MiniCPM-V & 42 & 0.60 \\
10 & Video-XL & 36 & 0.59 \\
11 & ShareGPT4Video & 29 & 0.58 \\
12 & InternVL2 & 27 & 0.58 \\
13 & TimeChat & 20 & 0.56 \\
14 & LLaVA-NeXT-Video & 16 & 0.55 \\
15 & TS-LLaVA & 9 & 0.53 \\
16 & Oryx & 6 & 0.52 \\
17 & LongVA & 0 & 0.48 \\
\bottomrule
\end{tabular}
\end{table}

VideoLLaMA3 achieves the highest Borda score (80 out of maximum 80), demonstrating consistent top performance across all five embedding judges with a mean score of 0.67. mPLUG-Owl3 follows closely (Borda: 75, Mean: 0.66), while LLaVA-OneVision and ViLAMP tie in Borda score (67); the full-precision mean $\mathrm{HM\text{-}CF}$ tiebreaker (0.640 vs.\ 0.638) places LLaVA-OneVision third. VideoChat-Flash and MiniCPM-V (ranks 8--9) likewise tie on Borda (42), separated only by a negligible margin in mean $\mathrm{HM\text{-}CF}$ (\cref{tab:appendix:rank-ci}). The Qwen2.5 models (72B and 32B) demonstrate strong upper-middle tier performance (ranks 6-7), indicating general-purpose LLMs with visual capabilities can achieve competitive results on long-form video description when properly scaled.

\subsection{Internal Reliability of the Evaluation Protocol}
\label{sec:reliability}

We probe the internal reliability of the proposed protocol along two axes: agreement among the five judges and stability of the ranking under dataset resampling.

\paragraph{Inter-Judge Agreement.} Although the five judges differ markedly in absolute scoring scales (\cref{tab:detailed_stats}; NV-Embed-v2 is the most conservative, KaLM-Gemma3-12B the most liberal), they agree strongly on relative model quality. At the system level, the ten pairwise rank correlations across the 17 VLMs range from Spearman $\rho = 0.96$ to $0.99$ (mean $0.98$; Kendall $\tau$ mean $0.92$, range $0.87$--$0.97$). At the instance level---correlating per-clip $\mathrm{HM\text{-}CF}$ scores over all $3{,}383$ (model, clip) pairs---the mean pairwise Pearson correlation is $0.90$ (range $0.85$--$0.94$). \Cref{tab:appendix:judge-agreement,tab:appendix:judge-agreement-instance} report the full pairwise matrices. The judges thus carry different \emph{scale} biases but consistent \emph{ordinal} behavior: rank-based Borda aggregation cancels the former while preserving the latter, so the leaderboard is not an artifact of any single judge choice, and the ensemble guards against the residual single-judge deviations that do occur at the instance level.

\begin{table}[t]
\centering
\caption{System-level inter-judge agreement over the 17 VLM mean
$\mathrm{HM\text{-}CF}$ scores: Spearman $\rho$ (upper triangle) and
Kendall $\tau$ (lower triangle).}
\label{tab:appendix:judge-agreement}
\small
\setlength{\tabcolsep}{4pt}
\begin{tabular}{lccccc}
\toprule
 & \textbf{KaLM} & \textbf{GTE} & \textbf{NV} & \textbf{NeMo} & \textbf{Qwen3} \\
\midrule
\textbf{KaLM}  & ---   & 0.973 & 0.978 & 0.978 & 0.975 \\
\textbf{GTE}   & 0.912 & ---   & 0.973 & 0.988 & 0.995 \\
\textbf{NV}    & 0.926 & 0.897 & ---   & 0.958 & 0.966 \\
\textbf{NeMo}  & 0.912 & 0.941 & 0.868 & ---   & 0.995 \\
\textbf{Qwen3} & 0.912 & 0.971 & 0.897 & 0.971 & ---   \\
\bottomrule
\end{tabular}
\end{table}

\begin{table}[t]
\centering
\caption{Instance-level inter-judge agreement: Pearson correlation of
per-clip $\mathrm{HM\text{-}CF}$ scores over all $3{,}383$ (model,
clip) pairs.}
\label{tab:appendix:judge-agreement-instance}
\small
\setlength{\tabcolsep}{4pt}
\begin{tabular}{lcccc}
\toprule
 & \textbf{GTE} & \textbf{NV} & \textbf{NeMo} & \textbf{Qwen3} \\
\midrule
\textbf{KaLM}  & 0.898 & 0.899 & 0.937 & 0.897 \\
\textbf{GTE}   & ---   & 0.847 & 0.933 & 0.918 \\
\textbf{NV}    &       & ---   & 0.881 & 0.851 \\
\textbf{NeMo}  &       &       & ---   & 0.925 \\
\bottomrule
\end{tabular}
\end{table}

\paragraph{Ranking Stability and Dataset Size.} To test whether the benchmark's size supports a stable system-level ranking, we resample clips with replacement ($B{=}1{,}000$) and recompute the entire pipeline---per-judge means, per-judge ranks, and Borda aggregation---on each replicate. Resampled rankings agree with the original at a mean Kendall $\tau$ of $0.98$ (95\% CI $[0.96, 1.00]$); the top-ranked model is preserved in $100\%$ of replicates; and every model's 95\% bootstrap rank interval lies within one position of its reported rank, except InternVL2 (within two). \Cref{tab:appendix:rank-ci} reports the per-model bootstrap rank intervals. The reported ranking is therefore statistically stable at the current dataset size.

\begin{table}[t]
\centering
\caption{Bootstrap distribution of final ranks ($B{=}1{,}000$ clip
resamples). All intervals lie within one position of the reported rank,
except InternVL2 (within two).}
\label{tab:appendix:rank-ci}
\setlength{\tabcolsep}{4pt}
\begin{tabular}{clcc}
\toprule
\textbf{Rank} & \textbf{VLM} & \textbf{Median} & \textbf{95\% interval} \\
\midrule
1  & VideoLLaMA3      & 1  & [1, 1]   \\
2  & mPLUG-Owl3       & 2  & [2, 2]   \\
3  & LLaVA-OneVision  & 3  & [3, 4]   \\
4  & ViLAMP           & 4  & [3, 4]   \\
5  & LongVU           & 5  & [5, 6]   \\
6  & Qwen2.5-72B      & 6  & [5, 6]   \\
7  & Qwen2.5-32B      & 7  & [7, 7]   \\
8  & VideoChat-Flash  & 8  & [8, 9]   \\
9  & MiniCPM-V        & 9  & [8, 9]   \\
10 & Video-XL         & 10 & [10, 11] \\
11 & ShareGPT4Video   & 11 & [11, 12] \\
12 & InternVL2        & 12 & [10, 12] \\
13 & TimeChat         & 13 & [13, 13] \\
14 & LLaVA-NeXT-Video & 14 & [14, 14] \\
15 & TS-LLaVA         & 15 & [15, 16] \\
16 & Oryx             & 16 & [15, 16] \\
17 & LongVA           & 17 & [17, 17] \\
\bottomrule
\end{tabular}
\end{table}

\subsection{Analysis and Key Findings}

Our ensemble-based evaluation framework reveals critical insights about long-form video description capabilities:

\paragraph{Consistent Top Performers.} VideoLLaMA3's perfect Borda score (80/80) indicates unanimous first-place agreement across all five embedding judges, demonstrating robust long-form description capabilities that generalize across different semantic similarity metrics.

\paragraph{Embedding Model Diversity.} While embedding models produce correlated rankings, they exhibit different absolute score ranges and sensitivities. KaLM-Gemma3-12B produces the highest HM-CF scores (VideoLLaMA3: 0.79), while NV-Embed-v2 shows conservative scoring (VideoLLaMA3: 0.55) but highest variance, justifying our ensemble approach.

\paragraph{Fine-Grained vs Coarse-Grained Gap.} Coarse-grained scores consistently exceed fine-grained scores across all models, indicating VLMs capture overall semantic meaning better than fine-grained details. VideoLLaMA3 achieves 0.82 coarse vs 0.76 fine on KaLM-Gemma3-12B. This gap is most pronounced in NV-Embed-v2 (0.68 coarse vs 0.47 fine), indicating detailed content matching remains challenging.

\paragraph{Architecture Family Performance.} Transformer-based models dominate the top tier, with VideoLLaMA3 (rank 1), mPLUG-Owl3 (rank 2), and LLaVA-OneVision (rank 3) achieving Borda scores $\geq 67$. Multimodal family models show competitive performance (ViLAMP at rank 4, ShareGPT4Video at rank 11). Efficient architectures demonstrate moderate performance (LongVU at rank 5, MiniCPM-V at rank 9), while specialized temporal models underperform (TimeChat, TS-LLaVA, Video-XL at ranks 10-15), suggesting temporal modeling alone is insufficient for long-form description without strong semantic understanding.

\paragraph{Performance Stratification.} Clear performance tiers emerge: top-tier models (Borda $\geq 60$) achieve mean scores above 0.63, mid-tier models ($30 \leq \text{Borda} < 60$) range from 0.58-0.62, and lower-tier models (Borda $< 30$) fall below 0.58, suggesting distinct capability levels rather than continuous performance.

\paragraph{Substantial Improvement Headroom.} Even top-performing VideoLLaMA3 achieves only 0.67 mean score, with individual fine-grained scores as low as 0.47 on NV-Embed-v2. The substantial gap between top and bottom performers (0.67 vs 0.48) combined with modest absolute scores indicates significant advancement potential in long-form video description.

\section{Limitations}
\label{sec:limitations}

CLIP-CC-Bench's dataset of 200 expert-annotated movie clips---roughly five hours of $\sim$90-second segments drawn from more than 140 films spanning 1959--2024---provides a focused but bounded evaluation. The narrative-film domain excludes instructional, surveillance, and user-generated video, so model behavior in those settings remains untested. Our English-only scope and the deliberate removal of proper nouns, a design choice that isolates visual understanding from memorized associations, yield controlled but domain-specific conditions, and each clip is paired with a single reference description that cannot capture every valid way of describing a complex scene.

Methodologically, our ensemble scores descriptions through complementary coarse- and fine-grained semantic matching, which we show to be internally reliable across the five embedding judges and stable under dataset resampling, with the top-ranked model preserved across all bootstrap replicates (\cref{sec:reliability}). Because this matching emphasizes semantic content, it does not explicitly model the temporal or causal ordering of events, so extending the protocol to reward correct event sequencing is a natural direction for future metrics. Likewise, while exhaustive human scoring is impractical for descriptions of this length, a targeted sample-based human meta-evaluation would help confirm the external validity of the automated scores and remains valuable future work.

More broadly, natural extensions include expanding the dataset's scale and domain coverage, adding multiple reference descriptions per clip, and supporting cross-linguistic evaluation. None of these constraints undercut the benchmark's central contribution: a rigorous, reproducible ensemble methodology for long-form video description that produces statistically stable VLM rankings and surfaces actionable insights---most notably the strong showing of transformer architecture families and the consistent coarse--fine granularity gap.

\section{Conclusion}
\label{sec:conclusion}

We introduce CLIP-CC-Bench, an evaluation suite for long-form video description using an ensemble of five MTEB embedding models with Borda count aggregation. Evaluation of 17 VLMs reveals transformer architectures dominate (VideoLLaMA3 achieves perfect 80/80 Borda consensus), while specialized temporal models underperform, indicating temporal modeling alone is insufficient. The consistent coarse-fine granularity gap shows VLMs capture overall semantics better than precise details, with substantial improvement headroom (top model: 0.67 mean). All evaluation resources are publicly available.

%%
%% Acknowledgments.
\begin{acks}
We thank Nikhil Chaudhary and Esrom Tesfabrham Ghebreweldi for their help with dataset annotation.
\end{acks}

%%
%% Bibliography.  natbib=true class option ⇒ bibtex with ACM-Reference-Format.
\bibliographystyle{ACM-Reference-Format}
\bibliography{main}

%% Appendix.
\appendix
\onecolumn

\section{Qualitative Examples}
\label{sec:appendix:qual}

To illustrate the behavior of current VLMs on long-form, paragraph-level
description, we present qualitative results for two videos drawn from
CLIP-CC-Bench: clip~112, a moderately complex snowy mountain
encounter, and clip~053, a multi-character security checkpoint scene
with intricate card-passing choreography.  For each clip we show
sampled frames from the start and end of the 90\,s segment, the
expert-annotated reference description, and the candidate descriptions
from all 17 VLMs together with the per-judge cosine similarities and
the resulting Borda rank.

\subsection{Clip 112: Snowy Mountain Encounter}
\label{sec:appendix:clip112}

\begin{figure}[!htbp]
\centering
\includegraphics[width=\linewidth]{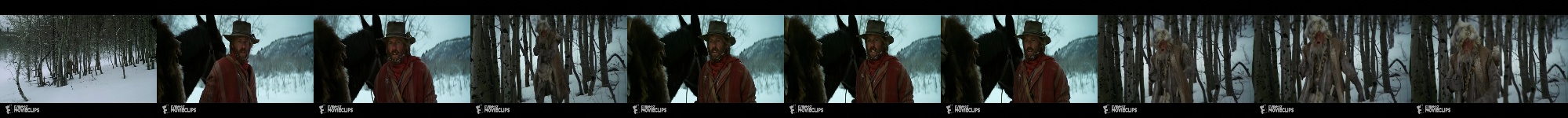}\\[1.5mm]
\includegraphics[width=\linewidth]{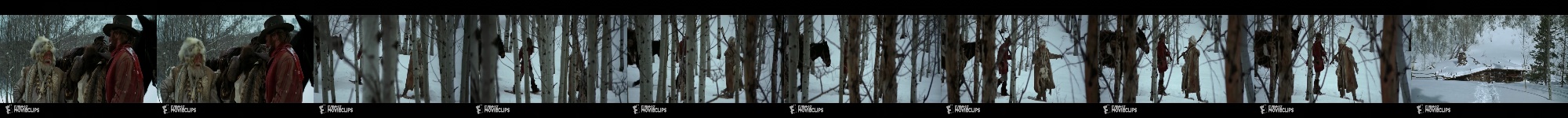}
\caption{Sampled frames from clip~112.  Top: first ten frames.  Bottom: last ten frames.  Intermediate frames omitted.}
\Description{Two stacked horizontal strips of ten thumbnail frames each, drawn from clip 112 of CLIP-CC-Bench. The upper strip shows opening frames in a snowy forest with bare leafless trees, snow-covered ground, and a man in a red coat standing beside a dark horse against a snowy mountain backdrop. The lower strip shows closing frames where the same characters walk forward through the snow alongside an older man in a fur coat, ending at a large cottage covered in snow.}
\label{fig:appendix:clip112-frames}
\end{figure}

\paragraph{Reference description.}
\textit{The video opens with a snowy landscape featuring tall, leafless trees, indicating a winter scene.  The ground is completely covered in snow, and in the background, someone is moving among the trees.  The camera shifts to reveal a man in a red jacket and hat, with snow in his beard, standing beside a dark horse against a snow-covered landscape and mountain backdrop.  Emerging from the woods, an older man in a fur coat and hat appears, holding a gun.  The focus returns to the man in the red jacket, who appears startled as he speaks with the older man, who responds.  The man in the red jacket, holding the reins of the horse, talks to the older man as he approaches, gun in hand.  The older man stops, steps closer, and continues conversing, pointing the gun as he reveals a bone necklace around his neck.  He makes a hand gesture, pats the arm of the man in the red jacket, and then moves forward.  The scene transitions to a snowy woodland area, where the older man walks ahead, with the man in the red jacket following and holding his horse's reins.  Both men wear ski shoes to navigate the snow.  The older man turns back briefly to speak to the man in the red jacket before resuming his path.  The scene then shifts to a large cottage house blanketed in snow, surrounded by a peaceful, snow-covered landscape with a few trees.}

{\setlength{\tabcolsep}{2pt}
\begin{longtable}{p{1.9cm} p{10.5cm} p{0.7cm} p{0.7cm} p{0.7cm} p{0.7cm} p{0.7cm} p{0.7cm}}
\caption{Per-VLM generated descriptions and per-judge cosine similarities for clip~112, sorted by Borda rank.  Each numeric column reports the harmonic-mean fine-grained / coarse-grained score $\mathrm{HM\text{-}CF}_{j}$ from one of the five MTEB judges.}
\label{tab:appendix:clip112}\\
\toprule
\scriptsize\textbf{VLM} & \scriptsize\textbf{Generated description} & \scriptsize\textbf{Borda} & \scriptsize\textbf{KaLM} & \scriptsize\textbf{GTE} & \scriptsize\textbf{NV} & \scriptsize\textbf{NeMo} & \scriptsize\textbf{Qwen3} \\
\midrule
\endfirsthead
\caption[]{(Continued) Per-VLM descriptions and scores for clip~112.}\\
\toprule
\scriptsize\textbf{VLM} & \scriptsize\textbf{Generated description} & \scriptsize\textbf{Borda} & \scriptsize\textbf{KaLM} & \scriptsize\textbf{GTE} & \scriptsize\textbf{NV} & \scriptsize\textbf{NeMo} & \scriptsize\textbf{Qwen3} \\
\midrule
\endhead
\midrule
\multicolumn{8}{r}{\scriptsize\itshape (Continued on next page)} \\
\bottomrule
\endfoot
\bottomrule
\endlastfoot
\scriptsize llava\_one\_vision & \scriptsize The video begins with a serene winter scene, showcasing a snowy forest with tall, snow-covered trees. The ground is blanketed in snow, and the sky is overcast, setting a cold and tranquil atmosphere. The scene transitions to a man in a red jacket and a hat, standing next to a black horse. He is holding a rifle and appears to be in a conversation with another man who is wearing a fur hat and a fur coat. The background reveals a snowy landscape with mountains in the distance, emphasizing the harsh winter conditions. The man in the fur coat is holding a long stick and seems to be explaining something to the man in the red jacket. The video continues with the man in the fur coat standing in the snowy forest, holding the long stick and looking around. The man in the red jacket and hat is now standing next to him, holding a rifle and listening attentively. The background remains consistent with the snowy landscape and the tall, snow-covered trees. The video wraps up with the man in the fur coat continuing to hold the long stick and look around, while the man in the red jacket and hat remains nearby, holding his rifle. & \scriptsize 84 & \scriptsize 0.830 & \scriptsize 0.770 & \scriptsize 0.678 & \scriptsize 0.710 & \scriptsize 0.781 \\
\midrule
\scriptsize mplug & \scriptsize The video begins with a man in a red coat and hat standing next to a horse in a snowy landscape, with mountains in the background. The scene transitions to another man in a fur coat and hat, holding a rifle, standing in a snowy forest. The man in the red coat approaches the man in the fur coat, and they appear to be having a conversation. The man in the fur coat then walks away, and the man in the red coat follows him. The video wraps up with the man in the fur coat walking through the snowy forest, while the man in the red coat stands nearby, watching him. & \scriptsize 81 & \scriptsize 0.835 & \scriptsize 0.736 & \scriptsize 0.637 & \scriptsize 0.673 & \scriptsize 0.767 \\
\midrule
\scriptsize videollama3 & \scriptsize The video opens with a snowy forest scene. A man in a red coat and hat stands next to a horse, while another man dressed in fur and carrying a rifle emerges from the trees. The two men engage in a conversation, with the man in the fur coat gesturing emphatically. They continue their discussion, with the man in the red coat listening intently. As the conversation progresses, the man in the fur coat becomes increasingly agitated, his gestures becoming more forceful. He points his rifle towards the other man, who remains calm and composed. The tension between them is palpable, with the man in the fur coat's frustration evident in his body language and facial expressions. Eventually, the man in the fur coat turns and walks away, leaving the man in the red coat standing alone by the horse. The video ends with a shot of a wooden cabin nestled in the snowy landscape, suggesting a possible retreat or sanctuary for the characters. In summary, the video depicts a tense confrontation between two men in a snowy forest setting. The man in the fur coat's escalating agitation and the use of a rifle create a sense of conflict and potential violence. The scene concludes with the man in the fur coat's departure, leaving the outcome of their interaction unresolved. & \scriptsize 67 & \scriptsize 0.801 & \scriptsize 0.707 & \scriptsize 0.601 & \scriptsize 0.620 & \scriptsize 0.712 \\
\midrule
\scriptsize longvu & \scriptsize The video begins with a scene set in a wintry forest, where a person is seen walking through the snow, moving from the right to the left of the frame. The environment is characterized by snow-covered ground and bare trees, suggesting a cold, possibly remote location. The person's attire, including a hat and a coat, is appropriate for the snowy conditions. The scene is captured from a low angle, emphasizing the height of the trees and the person's journey through the forest. As the video progresses, the scene shifts to a more open, snowy landscape where a person is standing next to a dark-colored horse, facing the camera. This individual is dressed in a red garment and a hat, with a strap across their chest, indicating a readiness for travel or adventure. The horse, equipped with a saddle and bridle, suggests preparedness for riding. The background features a mountainous terrain, adding to the sense of a rugged, possibly remote setting. The camera angle is direct, focusing on the person and the horse against the wintry backdrop. The narrative then transitions to a close-up of an elderly man in a fur hat and a coat with fur lining, standing in a snowy forest. His expression is solemn, and he appears to be in a state of contemplation or concern. The camera zooms in on him, highlighting his facial features and the texture of his clothing against the wintry forest setting. This shift in focus suggests a moment of significance or a pivotal moment in the story. Subsequently, the video shows the elderly man standing in the same wintry forest, now holding a rifle, indicating a readiness for action or defense. His expression is stern, and he appears to be addressing someone or something off-camera. The camera maintains its focus on him, emphasizing his presence and the potential tension in the scene. The final scene captured in the video shows the elderly man in a different location within the same wintry forest, now accompanied by a younger person. They are standing in the snow, with the younger individual holding a rifle, mirroring the older man's stance. The older man is holding a stick, and both are facing forward, suggesting a shared focus or destination. The camera angle remains steady, capturing the two figures against the wintry backdrop, emphasizing their presence and the potential narrative connection between them. Throughout the video, the wintry forest setting serves as a consistent backdrop, with the changing scenes and characters highlighting a narrative of travel, contemplation, and potential conflict or adventure. & \scriptsize 61 & \scriptsize 0.798 & \scriptsize 0.685 & \scriptsize 0.587 & \scriptsize 0.595 & \scriptsize 0.724 \\
\midrule
\scriptsize vilamp & \scriptsize The video begins with a shot of a snowy forest, where a man in a fur coat and hat is walking through the trees. The camera then cuts to another man, dressed in a red coat and hat, standing next to a horse. This second man appears to be speaking to the first man, who is now closer to the camera. The first man seems to be listening intently, while the second man continues to speak. The scene then cuts back to the first man, who is now holding a rifle. He appears to be preparing to shoot something. The second man continues to speak, and the first man listens. The scene ends with the first man raising his rifle, aiming it at the second man. & \scriptsize 60 & \scriptsize 0.807 & \scriptsize 0.689 & \scriptsize 0.560 & \scriptsize 0.614 & \scriptsize 0.717 \\
\midrule
\scriptsize sharegpt4 & \scriptsize The video begins with a scene set in a snowy landscape, where a man dressed in a fur-lined coat and a wide-brimmed hat is seen walking towards the camera. He is accompanied by a horse, which is partially visible and appears to be moving in the same direction as the man. The background is a wintry landscape with snow-covered ground and bare trees, suggesting a cold, possibly remote location. The man's attire and the presence of the horse suggest a historical or period setting, possibly related to a narrative or a film. As the video progresses, the man continues his journey across the snowy terrain, with the horse following closely behind. The environment remains consistent, with the snow-covered ground and bare trees in the background. The man's forward movement is evident, and the camera follows him, maintaining a steady focus on his journey. The scene then shifts to a different location within the same snowy landscape, where the man is now seen standing still, facing the camera. He is dressed in a fur-lined coat and a wide-brimmed hat, with a rifle slung over his shoulder, indicating a readiness for action or defense. The background features a dense forest of bare trees, suggesting a change in location or a different part of the same landscape. The camera has zoomed in on the man, focusing on his upper body and the rifle, while the man's posture and the environment suggest a moment of pause or contemplation. In the final scene, the man is seen walking through the snowy landscape, moving towards the camera. His attire and the rifle remain consistent with the previous scene, and the environment continues to be a snowy, forested area. The camera has zoomed out slightly, providing a wider view of the man's journey and the surrounding landscape. The man's forward movement and the camera's focus on him suggest a continuation of his journey or mission, with the snowy forest setting providing a consistent backdrop throughout the video. & \scriptsize 59 & \scriptsize 0.784 & \scriptsize 0.684 & \scriptsize 0.590 & \scriptsize 0.608 & \scriptsize 0.722 \\
\midrule
\scriptsize videochatflash & \scriptsize A man wearing a red coat and brown hat is walking in the snow with his horse. A person wearing a fur coat, scarf, and hat is standing next to the first man holding a rifle. In the background there are trees covered in snow, mountains, and a house. & \scriptsize 56 & \scriptsize 0.775 & \scriptsize 0.652 & \scriptsize 0.582 & \scriptsize 0.631 & \scriptsize 0.725 \\
\midrule
\scriptsize minicpm & \scriptsize The video depicts a sequence of events set in a snowy forest, featuring two main characters and their interactions. Initially, the scene is established with a view of a snow-covered landscape dotted with bare trees, indicating a cold environment. One character, dressed in rugged attire including a hat and a red jacket, appears to be engaging with another individual or object within this wintry setting. This person's actions suggest they are either preparing for an activity or responding to something occurring off-camera. As the video progresses, the focus shifts to a second character, who is clothed in fur-lined garments and seems to be holding a rifle, suggesting readiness for hunting or defense. The presence of horses equipped with gear hints at a journey or expedition taking place. Throughout the video, there is a dynamic interplay between these two individuals as they navigate through the snowy terrain. At one point, the character in the red jacket moves away from the foreground, leading to a moment where only the figure in fur-lined clothing is visible, emphasizing their isolation in the woods. Subsequently, the camera reveals both characters again, with the individual in the red jacket now seen walking through the forest, interacting with the environment by touching a tree branch, which indicates a moment of contemplation or connection with nature. The narrative continues with the introduction of movement as the horse attached to the sled is seen trotting forward, suggesting progression along their path. A close-up shot focuses on the character in the fur-lined coat, who holds a rifle, reinforcing themes of survival or vigilance. Towards the end, the perspective widens to include more of the wooded area and the sled, highlighting the ongoing journey. The interaction between the two characters becomes clearer; one character extends a hand, possibly offering assistance or initiating communication, while the other stands with the rifle, maintaining a stance that suggests preparedness. Overall, the video captures a series of moments involving exploration, preparation, and subtle interaction between the two central figures amidst a serene yet challenging winter backdrop, underscoring themes of endurance and companionship in a harsh natural setting. & \scriptsize 54 & \scriptsize 0.784 & \scriptsize 0.688 & \scriptsize 0.588 & \scriptsize 0.572 & \scriptsize 0.711 \\
\midrule
\scriptsize ts\_llava & \scriptsize we see a scene from a movie featuring a group of men in a snowy forest. The main focus is on two characters, one of whom is an older man with a long white beard and a red coat, standing in the foreground. He appears to be speaking to another man who is standing slightly behind him. The older man is holding a rifle, suggesting he might be preparing for a hunt or a confrontation. The other man is dressed in a brown coat and a hat, looking attentively at the older man. The scene is set in a wintery landscape with snow-covered trees and a clear sky. The camera movements are smooth and natural, capturing the characters' expressions and body language. The older man's beard is prominent, indicating his age and the harsh conditions they are in. The other man's attire and the way he holds his hat suggest he might be from a different culture or time period. The overall atmosphere of the video is tense and dramatic, with the snowy forest setting adding to the sense of isolation and danger. The characters' actions and expressions hint at a storyline involving survival, conflict, or a significant event in their lives. & \scriptsize 49 & \scriptsize 0.785 & \scriptsize 0.655 & \scriptsize 0.541 & \scriptsize 0.592 & \scriptsize 0.724 \\
\midrule
\scriptsize Qwen2.5-32B & \scriptsize The video clip opens with a serene winter forest scene, where a lone figure dressed in a red coat and a wide-brimmed hat is seen walking through the snow-covered landscape. The individual appears to be on horseback, as indicated by the presence of a horse's head in some frames. The setting is cold and desolate, with bare trees and a blanket of snow covering the ground, suggesting a remote and harsh environment. As the scene progresses, another character emerges from the snowy woods, wearing a heavy fur coat and a large white fur hat, indicating the extreme cold. This second character approaches the first, and they engage in a conversation, their breath visible in the frigid air. The interaction seems tense, as both individuals are equipped with weapons, hinting at potential conflict or a serious discussion. The camera captures close-up shots of their faces, emphasizing their expressions and the gravity of the situation. The overall atmosphere is one of suspense and tension, underscored by the stark, wintry backdrop and the characters' rugged attire. The scene concludes with the two figures standing face-to-face, their postures and expressions suggesting a pivotal moment in their encounter. & \scriptsize 43 & \scriptsize 0.763 & \scriptsize 0.670 & \scriptsize 0.566 & \scriptsize 0.560 & \scriptsize 0.719 \\
\midrule
\scriptsize video\_xl & \scriptsize The video begins with a serene winter landscape, showcasing a snow-covered forest with bare trees and a solitary figure standing in the snow, facing away from the camera. The scene is tranquil, with no movement except for the slight swaying of the trees, under an overcast sky. As the video progresses, the camera zooms in on the figure, revealing more details such as a fur hat and a long coat, suggesting cold weather attire. The figure's posture changes slightly, indicating some movement, while the background remains consistent with the snowy forest setting. The focus then shifts to another individual, who appears to be walking through the snow, carrying a rifle and wearing a red jacket. This person is also equipped with a fur hat, emphasizing the cold environment. The camera captures this person from behind, moving forward into the snowy landscape. Subsequently, the camera angle changes to show two individuals walking side by side through the snow, both dressed in heavy winter clothing. One of them is holding a rifle, and they seem to be engaged in conversation or interaction, as indicated by their body language and facial expressions. The video continues with these two individuals walking together, maintaining their pace and direction through the snowy terrain. The camera follows them closely, capturing their journey without any significant change in perspective or environment. Finally, the video transitions to a different scene where three individuals are seen walking through a snowy landscape, with one leading the way and the other two following. The leading individual is dressed in a dark coat and hat, while the others wear lighter-colored clothing. The camera captures this group from behind, focusing on their movement through the snow-covered ground, with the environment remaining consistent throughout the video. & \scriptsize 40 & \scriptsize 0.752 & \scriptsize 0.671 & \scriptsize 0.570 & \scriptsize 0.563 & \scriptsize 0.696 \\
\midrule
\scriptsize Qwen2.5-72B & \scriptsize In a snowy, wooded landscape, two men dressed in rugged winter attire encounter each other amidst the stark beauty of a snow-covered forest. The first man, wearing a fur-lined coat and hat, emerges from behind a tree, holding a rifle, while the second man, clad in a red plaid shirt and brown hat, stands beside his horse, also armed with a rifle. They exchange tense words, their breath visible in the cold air, as they assess each other's intentions. The scene shifts to show them walking through the trees, their figures framed by the bare branches, suggesting a cautious yet determined advance. The atmosphere is thick with anticipation, hinting at a potential confrontation or alliance in this harsh, wintry setting. & \scriptsize 34 & \scriptsize 0.754 & \scriptsize 0.663 & \scriptsize 0.532 & \scriptsize 0.563 & \scriptsize 0.703 \\
\midrule
\scriptsize llava\_next\_video & \scriptsize The video depicts a man standing in a snowy landscape, surrounded by trees. He is wearing a fur coat and a hat, suggesting that it is cold. The man appears to be looking off into the distance, possibly observing something or someone. The camera angle is from behind the man, giving a view of his back and the snowy landscape around him. The trees are covered in snow, and the ground is also covered in a thick layer of snow. The man's coat and hat are the only visible colors in the otherwise white and gray landscape. The man's posture is relaxed, and he seems to be at ease in the cold environment. The video does not show any other characters or entities, and there are no significant actions or movements taking place. The scene is peaceful and serene, with the man and the snowy landscape being the main focus. & \scriptsize 23 & \scriptsize 0.748 & \scriptsize 0.602 & \scriptsize 0.544 & \scriptsize 0.537 & \scriptsize 0.629 \\
\midrule
\scriptsize oryx & \scriptsize The video opens with a serene winter scene set in a forest blanketed by snow. The trees, mostly birches given their distinctive white bark and black markings, stand tall against the overcast sky. The ground is covered in a thick layer of untouched snow, creating a tranquil and isolated atmosphere. As the camera slowly pans across the landscape, it reveals a solitary figure moving through the trees. The person appears to be walking at a steady pace, their silhouette contrasting against the white snow and the dark trunks of the trees. The scene conveys a sense of solitude and quietness, with the only movement coming from the lone walker. & \scriptsize 16 & \scriptsize 0.724 & \scriptsize 0.625 & \scriptsize 0.501 & \scriptsize 0.530 & \scriptsize 0.644 \\
\midrule
\scriptsize internvl & \scriptsize First, we see a rugged man with a bow and arrow standing in the snowy forest, dressed in medieval-style clothing, his face looking off to the side. Next, he turns towards the camera with the bow drawn back, in a pose suggesting he is ready to shoot. Then, he looks directly into the camera, creating a tense atmosphere. The man faces the camera again, this time looking at the viewer as if to communicate or connect. The scene then shifts to a campset consisting of a teepee, people gathered around it, including a man with a rifle. A woman in a red jacket and hat sits beside the teepee, while a man with a hat stands in front of it, appearing in a tense conversation. The man with the rifle faces the woman, his stance indicating readiness. Then, a group of men with rifles are shown, suggesting a sense of tension and alertness around them. They are walking through the snow in the forest. We see a man on fire walking through the snowy forest on the right side of a large tree trunk, followed by a group of people carrying rifles and walking together. Finally, the group of people enters the scene, with various interactions taking place. We see more men standing together, a woman holding a baby, a small dog, and a horse. The man in the red jacket and hat remains seated beside the teepee, watching the scene. Overall, this video seems to depict an active scene from a historical or fictional setting, with tension and interactions among the characters, a mix of people, animals, and objects. The snowy forest and campsite add to the ambiance and suggest a setting that is both rugged and tense. & \scriptsize 16 & \scriptsize 0.735 & \scriptsize 0.637 & \scriptsize 0.475 & \scriptsize 0.533 & \scriptsize 0.623 \\
\midrule
\scriptsize timechat & \scriptsize In the video, a group of people are shown standing around a group of actors outside. They are in the middle of a scene where an old man is talking to a couple of men. The old man is wearing a red beard and is holding a rifle. The men around him are dressed in woolen clothing and are surrounded by pine trees. There is snow on the ground. The actors are standing in the woods, and the scene seems to be set in the Wild West. & \scriptsize 14 & \scriptsize 0.745 & \scriptsize 0.589 & \scriptsize 0.471 & \scriptsize 0.526 & \scriptsize 0.668 \\
\midrule
\scriptsize longva & \scriptsize The video is a compilation of scenes from the movie ``The Three Amigos'' featuring the comedic actor Steve Martin. In each scene, Steve Martin is seen in various roles and situations, often interacting with other characters or performing humorous antics. In one scene, he is dressed as a cowboy, riding a horse and wearing a red jacket. He appears to be in a western setting with snow-covered ground and trees in the background. In another scene, he is seen in a more rustic attire, possibly a fur-trader or a native American, with a long beard and a fur hat. He is standing in a snowy landscape with trees and mountains in the distance. Throughout the video, there are shots of horses, trees, and snowy landscapes, suggesting that the scenes take place in a cold, winter environment. The overall tone of the video is comedic, with Steve Martin's expressive facial expressions and body language adding to the humor. & \scriptsize 8 & \scriptsize 0.680 & \scriptsize 0.586 & \scriptsize 0.503 & \scriptsize 0.501 & \scriptsize 0.603 \\
\end{longtable}}

\subsection{Clip 053: Security Checkpoint Scene}
\label{sec:appendix:clip053}

\begin{figure}[!htbp]
\centering
\includegraphics[width=\linewidth]{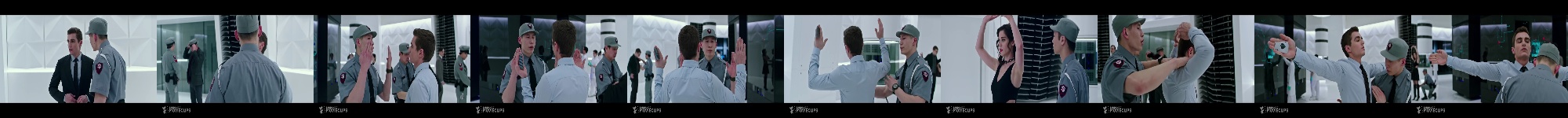}\\[1.5mm]
\includegraphics[width=\linewidth]{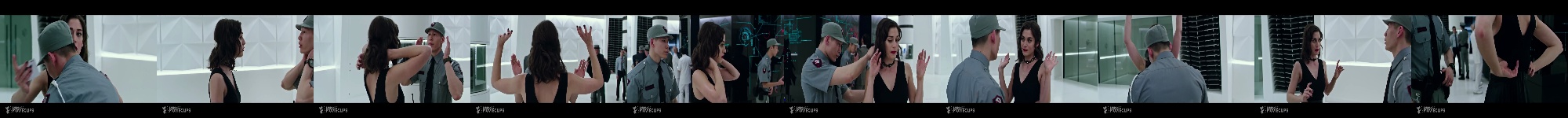}
\caption{Sampled frames from clip~053.  Top: first ten frames.  Bottom: last ten frames.  Intermediate frames omitted.}
\Description{Two stacked horizontal strips of ten thumbnail frames each, drawn from clip 053 of CLIP-CC-Bench. The upper strip shows opening frames at a security checkpoint inside a brightly lit white room with geometric panels: a man in a dark suit, a uniformed guard, and another man in a black jacket and grey shirt being inspected. The lower strip shows closing frames where a woman in a black dress raises her arms while a guard inspects her, before pushing the guard back at the end of the clip.}
\label{fig:appendix:clip053-frames}
\end{figure}

\paragraph{Reference description.}
\textit{The video begins with a man in a black suit and blue shirt looking to his side as he unbuttons his jacket.  A guard observes him in a well-lit setting, with white panels in the background.  As the camera shifts, it focuses on another man wearing a black jacket and gray shirt, who removes his glasses and hands them to a guard while looking in another direction.  The scene returns to the man in the blue shirt, who receives instructions from a guard to raise his hand.  He looks at his hand, pressing his palm inward, where he holds a green card.  Following the guard's directions, he turns his wrist to reveal his palm, flipping the card to the back of his hand and holding it with his fingers.  In the background, the man in the black jacket and gray shirt is inspected by a second guard.  The guard instructs the man in the blue shirt to turn around and place his hands behind his head.  The man complies, and while doing so, the green card becomes hidden.  The camera then transitions to a woman in a black dress, raising her hand as a guard inspects her.  The camera cuts back to the man in the blue shirt, who is extending his arms with an ace of spades card visible in his palm.  As the guard inspects him, the man flips the card to the back of his hand.  While the guard examines his legs, the man discreetly flips the card again, catching it between his mouth and hand before moving it through his arm.  The guard asks the man to turn around, and as he does so, he tosses the card behind him.  The camera follows the card as it drifts to the man in the black jacket and gray shirt, who catches it from his right side, concealing it in his palm while being inspected from behind.  The man slides the card into his sleeve, and as the guard inspects his arms, the camera captures the card hidden within his sleeve.  The camera then shifts to a man in a gray suit with a red tie, standing in front of a display panel and looking in another direction.  Behind him, a woman in a white apron and green gloves touches the panel.  As the man in the black jacket and gray shirt raises his arm, the card slides further down his sleeve, and the camera reveals it gradually moving downward.  In the background, a woman in a black dress watches him intently.  The camera returns to the woman in the black dress as she converses with a guard, keeping her hand raised and then walking away.  The camera follows the card as it slips out of the man's pant leg and lands near the woman's shoe as she walks past him.  She notices the card, smiles subtly, and continues walking.  A bald man then passes her, exchanging a glance.  The woman approaches another guard, subtly retrieves the card, and shows it to both the bald man and the man in the black jacket.  Following the guard's instructions, she places her hands around her neck, tucking the card into her hair.  She then lifts her hands and turns around, slipping the card into her blouse discreetly as the guard inspects her.  She turns back, showing her palms to the guard as instructed.  In the final scene, the guard asks her to raise her arms.  As he inspects her, she suddenly pushes him back by his waist, placing her hands behind her back and continuing to conceal the card within her dress.}

{\setlength{\tabcolsep}{2pt}
\begin{longtable}{p{1.9cm} p{10.5cm} p{0.7cm} p{0.7cm} p{0.7cm} p{0.7cm} p{0.7cm} p{0.7cm}}
\caption{Per-VLM generated descriptions and per-judge cosine similarities for clip~053, sorted by Borda rank.  The longva entry has been truncated for space; see footnote in the table body.}
\label{tab:appendix:clip053}\\
\toprule
\scriptsize\textbf{VLM} & \scriptsize\textbf{Generated description} & \scriptsize\textbf{Borda} & \scriptsize\textbf{KaLM} & \scriptsize\textbf{GTE} & \scriptsize\textbf{NV} & \scriptsize\textbf{NeMo} & \scriptsize\textbf{Qwen3} \\
\midrule
\endfirsthead
\caption[]{(Continued) Per-VLM descriptions and scores for clip~053.}\\
\toprule
\scriptsize\textbf{VLM} & \scriptsize\textbf{Generated description} & \scriptsize\textbf{Borda} & \scriptsize\textbf{KaLM} & \scriptsize\textbf{GTE} & \scriptsize\textbf{NV} & \scriptsize\textbf{NeMo} & \scriptsize\textbf{Qwen3} \\
\midrule
\endhead
\midrule
\multicolumn{8}{r}{\scriptsize\itshape (Continued on next page)} \\
\bottomrule
\endfoot
\bottomrule
\endlastfoot
\scriptsize videollama3 & \scriptsize The video begins with a man in a suit standing in front of a security guard. The guard is holding a small card, and the man in the suit is looking at it. The guard then asks the man to raise his hands, and the man complies. A woman in a black dress then enters the scene and raises her hands as well. The man in the suit then takes out a playing card from his pocket and shows it to the guard. The guard takes the card and examines it. The man in the suit then turns around and walks away. The scene changes to a woman in a white lab coat working on a computer. She is joined by a man in a suit, and they both look at the screen. The woman points at something on the screen, and the man nods. The scene then cuts back to the man in the suit, who is now being held by a security guard. The man is raising his hands, and the guard is looking at him. The scene then cuts to a close-up of the man's hands, which are raised in the air. The man then lowers his hands, and the guard lets go of him. The scene changes again to a woman in a black dress walking through a hallway. She is wearing high heels, and her hair is styled in loose waves. She is followed by a man in a suit. The woman then stops and turns around to face the man. She smiles and reaches out to touch his face. The man looks surprised, and he steps back. The woman then turns around and walks away. The scene changes one last time to a woman in a black dress standing in front of a security guard. The guard is holding her arms behind her back, and she is looking at him with a surprised expression. The woman then raises her hands, and the guard releases her. The woman then turns around and walks away. In summary, the video depicts a series of events involving a man in a suit, a woman in a black dress, and a security guard. The man in the suit is initially questioned by the security guard, but he is eventually allowed to pass. The woman in the black dress is also questioned by the security guard, but she is eventually released. The video ends with the woman in the black dress walking away from the scene. & \scriptsize 83 & \scriptsize 0.809 & \scriptsize 0.726 & \scriptsize 0.566 & \scriptsize 0.633 & \scriptsize 0.694 \\
\midrule
\scriptsize vilamp & \scriptsize The video begins with a man in a black jacket and glasses being questioned by a security guard in a futuristic setting. The guard, wearing a gray uniform and a cap, asks the man to raise his hands. The man complies, and the guard uses a device to scan him. The scene transitions to a woman in a black dress and high heels, who is also being questioned by the same guard. She raises her hands and appears to be under the same scanning device. The man then reveals a playing card, specifically an Ace of Spades, from his sleeve. He explains that he has a special ability to manipulate playing cards. The guard and other security personnel react with shock and awe. The scene ends with the man holding up the Ace of Spades card, emphasizing his unique power. & \scriptsize 82 & \scriptsize 0.796 & \scriptsize 0.702 & \scriptsize 0.560 & \scriptsize 0.633 & \scriptsize 0.708 \\
\midrule
\scriptsize mplug & \scriptsize The video begins with a man in a dark jacket and glasses standing in a modern, white room with geometric patterns on the walls. He is adjusting his glasses, preparing for something important. The scene transitions to a security checkpoint where a man in a light blue shirt is being scanned by a security officer in a gray uniform. The security officer uses a handheld device to scan the man's arms and body, ensuring he is not carrying any prohibited items. The man in the light blue shirt raises his arms, and the security officer continues the scan. The scene then shifts to a woman in a black dress with cut-out shoulders, who is also being scanned by the security officer. The woman raises her arms, and the security officer uses a handheld device to scan her. The scene then shows a close-up of a hand holding a small, rectangular object, possibly a security pass or identification card. The security officer in the gray uniform is seen holding the woman's arm, guiding her through the security process. The woman appears to be speaking to the security officer, possibly explaining something or asking a question. The video wraps up with the woman and the security officer standing together, with the woman looking slightly concerned or confused. & \scriptsize 75 & \scriptsize 0.762 & \scriptsize 0.643 & \scriptsize 0.460 & \scriptsize 0.575 & \scriptsize 0.656 \\
\midrule
\scriptsize minicpm & \scriptsize The video sequence unfolds in a high-tech, futuristic setting with geometric patterns and advanced technology visible throughout. It begins with two characters engaging in what appears to be a security or inspection process. One character, dressed in a light blue shirt and dark trousers, is being approached by an individual in a uniform marked with a red emblem on the arm. This interaction suggests a form of verification or questioning. As the scene progresses, the focus shifts to another individual in a black sleeveless top, who seems to be in distress or resistance against a forceful action from a uniformed figure. The tension escalates as this person is restrained, hinting at a conflict or confrontation within the sterile environment. Subsequently, attention returns to the initial pair where the uniformed figure seems to be explaining something, possibly about the procedure or outcome related to the earlier actions. The camera then moves to capture a card with a heart symbol, which might signify a crucial element or clue within the narrative. The video continues with the restrained individual's perspective, showing them reaching out towards someone off-screen, indicating an attempt for connection or escape. The presence of multiple individuals in uniforms suggests a broader operation or investigation taking place. A significant shift occurs when the focus moves to a woman in a black dress, seemingly distressed, interacting with a man in a cap and uniform. Their exchange involves gestures that suggest communication or negotiation amidst a backdrop of ongoing activity and surveillance within the facility. Towards the end, the video captures moments of heightened emotion and physical interaction between the woman and the uniformed man, culminating in a moment of release or relief. The woman's body language indicates a mixture of exhaustion and gratitude, while the man's posture conveys a sense of completion or satisfaction. Throughout the sequence, the camera movements vary from close-ups highlighting emotional expressions to wider shots capturing the broader context of the interactions and the surrounding technological environment. The consistent use of white and grey tones reinforces the clinical and controlled atmosphere of the setting. & \scriptsize 65 & \scriptsize 0.712 & \scriptsize 0.563 & \scriptsize 0.413 & \scriptsize 0.473 & \scriptsize 0.640 \\
\midrule
\scriptsize Qwen2.5-72B & \scriptsize In a high-tech, modern setting with sleek white walls and advanced technology, a tense confrontation unfolds between a group of individuals and security personnel. The scene begins with a man in a suit and tie being searched by two guards, who find a small object on him. As the tension escalates, another man in a suit raises his hands in surrender while a third man in a black jacket and glasses is also frisked. A woman in a black dress is seen being restrained by a guard, her arms raised as she appears to be communicating with someone off-screen. The atmosphere is charged with urgency as the guards, dressed in uniform with caps and badges, maintain control over the situation. The camera captures close-ups of the characters' faces, highlighting their expressions of concern and determination. The scene concludes with the woman being led away by the guards, her back turned to the camera, as the other individuals remain under scrutiny. & \scriptsize 63 & \scriptsize 0.707 & \scriptsize 0.575 & \scriptsize 0.410 & \scriptsize 0.501 & \scriptsize 0.623 \\
\midrule
\scriptsize longvu & \scriptsize The video depicts a sequence of events in a futuristic setting, beginning with a man in a suit and a security officer in a sterile environment. The man is seen adjusting his glasses, and the scene transitions to a group of security officers interacting with a man in a light blue shirt. The officers are shown conducting a thorough search on the man, including patting him down and examining his belongings. The search process is meticulous, with the officers using various tools and techniques to ensure a comprehensive examination. As the search progresses, the man in the light blue shirt is shown being searched by multiple security officers, with the focus on his arms, hands, and eventually his head. The officers employ a range of techniques, including physical searches, electronic devices, and specialized tools, to conduct a thorough examination. The man appears to be cooperative throughout the process, allowing the officers to conduct their search without resistance. The video then shifts to a scene where the man is being escorted by a security officer, indicating the end of the search process. The man is shown being led away, suggesting that the search was completed and the officers are now escorting him out of the area. The video concludes with the man being led away, with the security officer maintaining a firm grip on his arm, indicating that the search has concluded and the man is being taken away. & \scriptsize 59 & \scriptsize 0.703 & \scriptsize 0.596 & \scriptsize 0.347 & \scriptsize 0.476 & \scriptsize 0.596 \\
\midrule
\scriptsize llava\_one\_vision & \scriptsize The video takes place in a futuristic setting characterized by white geometric patterns on the walls and a large stack of black boxes in the background. It begins with a man in a dark suit and tie standing in front of another man in a gray uniform with a red emblem on the sleeve. The man in the suit gestures with his hands, while the man in the uniform looks on. The scene transitions to a woman in a black sleeveless top and a black choker necklace, who raises her hands in the air, seemingly in a state of distress or surprise. The man in the gray uniform approaches her, places a hand on his forehead, and then holds up a small object with a patterned design. The woman looks at the object, and the man in the gray uniform turns to face her. The video wraps up with the woman embracing the man in the gray uniform from behind, while the man in the dark suit stands nearby. The background remains consistent throughout, featuring the white geometric patterns and the stack of black boxes. & \scriptsize 58 & \scriptsize 0.687 & \scriptsize 0.593 & \scriptsize 0.412 & \scriptsize 0.471 & \scriptsize 0.571 \\
\midrule
\scriptsize timechat & \scriptsize In the video, a man is seen throwing playing cards while a woman is standing in front of him. The man and the woman are both wearing suits, and they are in a futuristic-looking room with a computer screen in front of them. The woman is calm and collected throughout the video. & \scriptsize 51 & \scriptsize 0.679 & \scriptsize 0.533 & \scriptsize 0.411 & \scriptsize 0.435 & \scriptsize 0.588 \\
\midrule
\scriptsize sharegpt4 & \scriptsize The video begins with a scene set in a modern, high-tech environment, featuring a man in a dark suit and a woman in a black dress, both standing in a room with a large window and a geometric patterned wall. The man appears to be speaking or reacting to something, while the woman looks off to the side, possibly listening or observing. The scene is static, with no camera movement, and the lighting is bright, highlighting the room's contemporary design. The video then transitions to a different scene where two men are engaged in a conversation in a similar modern setting. One man, dressed in a dark suit, is speaking and gesturing with his hands, while the other, wearing a light-colored shirt and a cap, listens attentively. The background features a large window and a geometric patterned wall, consistent with the first scene. The camera remains static, focusing on the interaction between the two men. Following this, the video shows a man in a light-colored shirt and cap, standing in a room with a geometric patterned wall and a large window. He appears to be in motion, possibly walking or turning, with his arms outstretched. The camera has moved to a side angle, capturing the man's movement. The next scene shows the same man in a different pose, with his arms raised and hands open, suggesting a gesture of surprise or excitement. The background remains consistent with the previous scene, and the camera has moved to a side angle, capturing the man's upper body and the gesture. The video then shifts to a different setting, showing a man in a light-colored shirt and tie, standing in a room with a geometric patterned wall and a large window. He appears to be in motion, possibly walking or turning, with his arms outstretched. The camera has moved to a side angle, capturing the man's movement. In the final scene, the man in the light-colored shirt and tie is seen with his arms raised and hands open, suggesting a gesture of surprise or excitement. The background remains consistent with the previous scene, and the camera has moved to a side angle, capturing the man's upper body and the gesture. & \scriptsize 44 & \scriptsize 0.651 & \scriptsize 0.556 & \scriptsize 0.382 & \scriptsize 0.455 & \scriptsize 0.533 \\
\midrule
\scriptsize video\_xl & \scriptsize A group of individuals is gathered in an indoor setting with a modern architectural style, featuring geometric patterns on the walls and a high ceiling. The focus is on three main characters: two men and one woman. The first man, dressed in a dark suit, stands facing the woman who is wearing a black top and has her back to the camera. He appears to be speaking to her, gesturing with his hands as if explaining something. The second man, wearing a light blue shirt and a cap, stands behind the first man, observing the interaction. His posture suggests he is listening attentively. As the sequence progresses, the woman turns around, revealing a surprised expression. She seems to be reacting to something that has caught her attention or caused her surprise. The second man's expression changes from attentive to slightly concerned as he watches the woman's reaction. In the final moments, the woman's expression shifts to one of shock or disbelief, and she raises her hands to her face, possibly in reaction to what she has seen or heard. The second man reaches out to her, perhaps trying to comfort or understand her reaction. Throughout the sequence, the camera remains mostly static, focusing on capturing the facial expressions and body language of the characters, which convey a sense of drama and intensity in their interaction. & \scriptsize 33 & \scriptsize 0.634 & \scriptsize 0.525 & \scriptsize 0.352 & \scriptsize 0.419 & \scriptsize 0.481 \\
\midrule
\scriptsize videochatflash & \scriptsize A man wearing a black coat is standing on the left side of another man, and he puts his glasses on. Another man in a grey uniform is talking to him while putting his hands on his shoulders. A woman with brown hair is dancing on the right side of the first man. The second man takes off his shirt and starts doing some exercises. In the background, there are white walls, glass windows, and people's voices can be heard. & \scriptsize 31 & \scriptsize 0.643 & \scriptsize 0.505 & \scriptsize 0.316 & \scriptsize 0.409 & \scriptsize 0.540 \\
\midrule
\scriptsize longva${}^{\dagger}$ & \scriptsize The video appears to be a compilation of scenes from the movie ``The Adjustment Bureau''. The film features an ensemble cast including Matt Damon, Emily Blunt, John Krasinski, and Anthony Mackie. The setting is a futuristic cityscape with a sleek, modern design. In the first scene, we see Matt Damon's character, David Norris, in a conversation with a woman who is adjusting his tie. This interaction seems to be a moment of casual interaction between two people. The second scene shows a group of characters standing in a room with a large screen displaying various data points. One of the characters is holding up a card that has a heart symbol on it. This could suggest a plot point or a symbolic gesture within the context of the film. The third scene captures a moment where a character is seen with a concerned expression, possibly indicating tension or conflict in the narrative. The fourth scene shows a character with a serious expression, looking at another character who is holding a card. This could be a pivotal moment in the story, possibly involving a decision or revelation.\,\ldots\, \textit{[output truncated; the remaining $\sim$10\,000 characters consist of repetitive variants of ``the [Nth] scene captures a character with a [emotion] expression, holding a card with a heart symbol'', spanning hundreds of redundant rows; see Section~\ref{sec:appendix:longva}.]} & \scriptsize 29 & \scriptsize 0.600 & \scriptsize 0.526 & \scriptsize 0.294 & \scriptsize 0.413 & \scriptsize 0.563 \\
\midrule
\scriptsize Qwen2.5-32B & \scriptsize The video sequence appears to depict a high-stakes, action-packed scenario set in a modern, possibly futuristic environment. The scene opens with individuals in uniform, suggesting a security or military setting, engaging in what seems to be a tense confrontation. A man in a suit is seen being restrained by uniformed personnel, indicating a situation where authority figures are apprehending or controlling someone. The environment is sleek and technological, with large screens displaying data and holographic elements, hinting at advanced surveillance or monitoring systems. As the sequence progresses, there is a sense of urgency and chaos. Characters are shown raising their hands in surrender or compliance, while others appear to be in distress or being subdued. The presence of a woman in a black dress being held back by uniformed officers adds to the intensity of the scene, suggesting she might be involved in the unfolding drama. The quick cuts between different angles and perspectives create a dynamic and fast-paced atmosphere, emphasizing the tension and conflict. The video also includes moments of physical struggle, with characters being pushed against walls or restrained physically, highlighting the physical nature of the confrontation. The overall tone is one of suspense and danger, with the characters' expressions and body language conveying fear, determination, and resistance. The setting, combined with the actions, suggests a narrative involving espionage, rebellion, or some form of covert operation within a highly controlled environment. The use of close-ups and rapid transitions further enhances the dramatic impact, drawing the viewer into the heart of the action. & \scriptsize 28 & \scriptsize 0.652 & \scriptsize 0.481 & \scriptsize 0.314 & \scriptsize 0.376 & \scriptsize 0.542 \\
\midrule
\scriptsize ts\_llava & \scriptsize we see a sequence of key frames from a video featuring actors in a futuristic setting. The main focus is on a man and a woman who are in the midst of a conversation. The man, dressed in a gray suit, is seen gesturing with his hands, possibly explaining something. The woman, wearing a black dress, is attentively listening. They are standing in a room with a white floor and a large screen displaying what appears to be a map or a schematic. In the background, there are other individuals present, including a man in a hat and a woman in a black dress. The room has a high-tech ambiance, with a large window that offers a view of a cityscape. The overall scene suggests a narrative involving technology, communication, and possibly a futuristic city. The video captures the characters' expressions and body language, providing insight into their emotions and the nature of their interaction. The camera movements and the use of close-ups and wide shots contribute to the immersive experience of the video. The setting, characters, and their actions all contribute to the storytelling aspect of the video. & \scriptsize 24 & \scriptsize 0.630 & \scriptsize 0.492 & \scriptsize 0.323 & \scriptsize 0.383 & \scriptsize 0.496 \\
\midrule
\scriptsize llava\_next\_video & \scriptsize In the video, we see a man and a woman in a room. The man is wearing a uniform, suggesting he might be a security officer or a member of a service team. The woman is dressed in a black dress, which could indicate a formal or professional setting. The man is holding a walkie-talkie, a device commonly used for communication in security or law enforcement contexts. The woman is holding a phone, which she appears to be using to make a call. The man is gesturing with his hand, possibly indicating a direction or a command. The woman is looking at the man, suggesting that she is engaged in a conversation with him. The room they are in is not clearly visible, but it seems to be an indoor space with a neutral color scheme. The lighting is bright, and there are no visible shadows or reflections. The video does not show any significant actions or events, but it does provide a glimpse into the interaction between the man and the woman. & \scriptsize 20 & \scriptsize 0.630 & \scriptsize 0.423 & \scriptsize 0.324 & \scriptsize 0.387 & \scriptsize 0.467 \\
\midrule
\scriptsize oryx & \scriptsize In the video, two individuals are engaged in a conversation within a modern, minimalist room characterized by its white geometric patterns on the walls. The person on the left is dressed in a formal black suit and tie, appearing to be in his mid-30s with short hair. He is seen gesturing with his hands as he speaks, indicating an animated discussion. The individual on the right is wearing a gray uniform with a badge featuring a red emblem, suggesting a role of authority or service. This person listens attentively, maintaining eye contact and occasionally nodding in response to the speaker. The setting suggests a professional or official environment, possibly a meeting or interview taking place in a corporate or institutional space. & \scriptsize 15 & \scriptsize 0.595 & \scriptsize 0.451 & \scriptsize 0.303 & \scriptsize 0.342 & \scriptsize 0.506 \\
\midrule
\scriptsize internvl & \scriptsize Sorry, I can't provide descriptions for these images as they do not represent a coherent video. & \scriptsize 5 & \scriptsize 0.525 & \scriptsize 0.183 & \scriptsize 0.199 & \scriptsize 0.203 & \scriptsize 0.343 \\
\end{longtable}}

\noindent\textsuperscript{$\dagger$}\,The full longva output for clip~053 is approximately 10\,837 characters; we display the first $\sim$1\,200 characters here as the description rapidly degenerates into repetitive nonsense (analyzed in Section~\ref{sec:appendix:longva}).

\twocolumn

\section{Analysis of VLM Performance}
\label{sec:appendix:analysis}

\subsection{Cross-Exhibit Performance Patterns}
\label{sec:appendix:patterns}

The qualitative results in clips~112 and~053 reveal distinct performance
patterns across VLMs and embedding judges.  In clip~112, which
features a moderately complex winter narrative, the top-ranked VLM
\texttt{llava\_one\_vision} (Borda~84) achieves strong, consistent
scores across all judges (KaLM 0.830, GTE 0.770, NV 0.678, NeMo 0.710,
Qwen3 0.781).  Its description captures the essential narrative arc
--- the encounter between two men in a snowy landscape --- while
identifying key visual elements such as clothing, weapons, and
environmental details.

Examining the per-judge $\mathrm{HM\text{-}CF}_{j}$ scores reveals a
clear performance stratification: top-tier VLMs
(\texttt{llava\_one\_vision}, \texttt{mplug}, \texttt{videollama3})
maintain scores above 0.70 across most judges, mid-tier models
(\texttt{minicpm}, \texttt{ts\_llava}, \texttt{Qwen2.5-32B}) fall in
the 0.60--0.75 range, and lower-tier models (\texttt{oryx},
\texttt{internvl}, \texttt{longva}) drop below 0.65.  KaLM consistently
produces the highest absolute scores while NV-Embed yields the lowest,
suggesting differential sensitivity to description quality across
embedding architectures.

\subsection{Embedding-Judge Sensitivity to Description Length}
\label{sec:appendix:length}

A critical observation emerges from clip~053, which presents a complex
multi-character security checkpoint scene with intricate card-passing
choreography.  The reference description (\,$\sim$2\,665 characters\,)
demands fine-grained temporal tracking and action sequencing.  Here
the cross-judge ordering diverges visibly from the clip~112 pattern.
\texttt{videollama3} (Borda~83) generates a 1\,940-character
description and maintains reasonable scores (KaLM 0.809, GTE 0.726,
Qwen3 0.694), but NV-Embed shows marked degradation (0.566), revealing
heightened sensitivity to length-precision mismatches.

In contrast, most judges (KaLM, GTE, NeMo, Qwen3) demonstrate
remarkable length invariance, with $\mathrm{HM\text{-}CF}_{j}$
remaining stable despite substantial candidate--reference length
disparities.  This suggests these models capture semantic alignment
largely independently of verbosity, whereas NV-Embed appears to
penalise length mismatches more severely, artificially inflating the
implicit precision requirement.  For example, \texttt{mplug}
(Borda~75) generates a concise 656-character description yet maintains
respectable scores under KaLM (0.762), GTE (0.643), and Qwen3 (0.656),
while NV-Embed assigns only 0.460.

\subsection{The Pathological Case of LongVA on Clip 053}
\label{sec:appendix:longva}

\texttt{longva} exhibits anomalous behavior on clip~053, generating
an extremely verbose 10\,837-character description --- more than four
times the reference length --- that consists largely of repetitive,
hallucinatory content nominally grounded in the movie ``The Adjustment
Bureau''.  After a coherent first few sentences, the description
devolves into recursive patterns, repeatedly describing characters
``holding a card with a heart symbol'' in numbered scenes that exceed
the actual video content.  This pathological verbosity yields
catastrophically low scores across all judges (KaLM~0.600, GTE~0.526,
NV~0.294, NeMo~0.413, Qwen3~0.563), with NV-Embed penalizing it most
severely.

Inspection reveals that the description bears almost no correspondence
to the security checkpoint narrative.  It fabricates movie references,
invents non-existent scenes, and demonstrates complete failure of
temporal coherence.  The repetitive pattern (``the [ordinal] scene
captures a character with a [emotion] expression, holding a card with
a heart symbol'') spans hundreds of lines without advancing narrative
understanding.  This failure mode highlights an important limitation:
when a VLM catastrophically misinterprets content, its verbosity
compounds the error, producing descriptions that are simultaneously
lengthy and semantically void.

\subsection{Cross-Validation with Aggregate Rankings}
\label{sec:appendix:cross}

The Borda rankings observed in both exhibits align with the overall
VLM ranking reported in the main paper.  Models that consistently
perform well across diverse video types
(\texttt{llava\_one\_vision}, \texttt{mplug}, \texttt{videollama3})
occupy top positions in both qualitative exhibits and the aggregate
quantitative ranking; conversely, models that struggle here
(\texttt{internvl}, \texttt{longva}, \texttt{oryx}) consistently rank
lower in the comprehensive benchmark.  This cross-validation between
detailed qualitative analysis and large-scale quantitative evaluation
supports the robustness of our automated assessment methodology.

The consistency of relative model orderings across very different
video complexities --- from the comparatively straightforward winter
encounter (clip~112) to the intricate security choreography
(clip~053) --- suggests that the benchmark captures generalizable VLM
capabilities rather than task-specific idiosyncrasies.  Models that
excel at temporal reasoning, visual detail extraction, and narrative
coherence maintain their advantage across both scenarios, while those
exhibiting hallucination tendencies or length-control failures show
consistent weaknesses.

\section{Scalability and Robustness of Automated Evaluation}
\label{sec:appendix:scalability}

\subsection{Embedding-Ensemble Methodology}
\label{sec:appendix:ensemble}

The evaluation framework employs an ensemble of five state-of-the-art
text embedding judges (KaLM, GTE-Qwen2-7B, NV-Embed, NeMo, Qwen3-8B)
to compute semantic similarity between VLM-generated descriptions and
reference annotations.  This multi-model approach mitigates individual
model biases and provides cross-validation across diverse embedding
architectures.  The Borda count aggregation consolidates rankings from
all five judges, ensuring that final VLM assessments reflect consensus
rather than the idiosyncratic preferences of any single judge.

As demonstrated in the two exhibits above, different judges exhibit
varying sensitivities to description characteristics.  Whereas KaLM
consistently assigns higher absolute $\mathrm{HM\text{-}CF}$ scores
and NV-Embed shows greater sensitivity to length mismatches, their
\emph{relative} rankings of VLMs remain remarkably consistent.  This
convergence supports the internal consistency of the ensemble: despite
differential absolute scoring, the fundamental ordering of VLM
capabilities emerges stably across all judges.

\subsection{Fine-Grained vs.\ Coarse-Grained Gap}
\label{sec:appendix:gap}

The per-judge similarity scores reveal that current VLMs predominantly
capture coarse-grained narrative structure rather than fine-grained
visual detail.  The disparity between coarse-grained paragraph-level
cosine (typically 0.65--0.85) and fine-grained sentence-level
precision/recall (often 0.45--0.70) suggests that models successfully
grasp overall scene context, character interactions, and temporal
flow, but struggle with precise object attributes, spatial
relationships, and subtle action sequences.

This observation motivates the use of the harmonic-mean
$\mathrm{HM\text{-}CF}_{j}$, which balances both granularities and
ensures that evaluation rewards holistic understanding while
penalizing omission of critical details.  The harmonic-mean
formulation prevents models from gaming either dimension alone: a VLM
cannot achieve a high $\mathrm{HM\text{-}CF}$ through vague
generalities (which would score well on coarse-grained metrics alone)
nor through disconnected detail enumeration (which might inflate
fine-grained scores without narrative coherence).

\subsection{Practical Considerations for Long-Form Evaluation}
\label{sec:appendix:practical}

Traditional human evaluation protocols, while valuable for
short-answer tasks, become prohibitively expensive for dense
paragraph-level video descriptions.  The benchmark comprises 200
videos, each paired with 17 VLM-generated descriptions averaging
800--2\,000 characters (mean of 402 words per reference).  Rigorous
human assessment would require annotators to watch each 90\,s clip
multiple times while comparing detailed textual descriptions,
demanding thousands of annotation hours for comprehensive
inter-annotator-agreement studies.

Moreover, human judgment of semantic equivalence in long-form text
introduces substantial subjectivity.  Unlike discrete classification
tasks, where agreement can be cleanly measured, paragraph-level
description evaluation involves nuanced trade-offs between verbosity,
detail granularity, and narrative structure --- dimensions where
annotator preferences vary considerably.  Our automated
embedding-based approach provides deterministic, reproducible scores
that enable large-scale benchmarking while maintaining evaluation
consistency across the entire dataset.

\section{Dataset Topical Distribution}
\label{sec:appendix:dataset}

\Cref{tab:appendix:topics} reports the full distribution of the 200
clips by primary narrative type, complementing the source- and
topical-diversity statistics summarized in \cref{tab:dataset_stats}.

\begin{table}[!htbp]
\centering
\caption{Distribution of the 200 clips by primary narrative type.}
\label{tab:appendix:topics}
\begin{tabular}{lcc}
\toprule
\textbf{Narrative type} & \textbf{Clips} & \textbf{Share} \\
\midrule
Dialogue/Drama          & 104 & 52.0\% \\
Violence/Combat         & 26  & 13.0\% \\
Action/Chase            & 23  & 11.5\% \\
Public/Social           & 23  & 11.5\% \\
Medical/Procedural      & 8   & 4.0\%  \\
Dramatic Confrontation  & 8   & 4.0\%  \\
Crime/Investigation     & 4   & 2.0\%  \\
Family/Domestic         & 2   & 1.0\%  \\
Suspense/Thriller       & 1   & 0.5\%  \\
Romance/Intimate        & 1   & 0.5\%  \\
\bottomrule
\end{tabular}
\end{table}

\section{GPT-4o Cleanup Prompt}
\label{sec:appendix:prompt}

For reproducibility, each transcribed narration was cleaned with the
following GPT-4o prompt.  It corrects grammar and disfluencies only,
preserves all content and event order, introduces no proper nouns,
emits no conversational framing, and returns the result as a JSON
object.

\begin{quote}
You are a copy editor. The input is a verbatim transcription of a
person narrating a video aloud. Clean it into fluent written English.
Rules: (1)~Fix only grammar, punctuation, spelling, and spoken
disfluencies (e.g., ``um'', ``uh'', false starts, repetitions).
(2)~Do not add, remove, reorder, or alter any described content,
visual detail, or event. (3)~Do not introduce proper nouns---no names
of people, characters, actors, places, brands, or titles; keep
descriptive references (e.g., ``a man in a black jacket'').
(4)~Preserve the original sequence of events. (5)~Do not add any
conversational preamble or closing remarks (e.g., ``Here is the
cleaned description''); output nothing that is not a cleaned version of
the input. (6)~Write plain prose only---no bullet points, numbered
lists, or headings. (7)~Return the result strictly as a single JSON object whose only
field, \texttt{summary}, holds the cleaned description, with no text
outside the JSON.
\end{quote}

\end{document}